\documentclass[10pt]{article} 

\usepackage[utf8]{inputenc}
\usepackage{amsmath}
\usepackage{amsfonts}
\usepackage{amssymb}
\usepackage{mathtools}
\usepackage{listings}
\usepackage{mathrsfs}
\usepackage{times}
\usepackage{float}
\usepackage{color}
\usepackage{setspace}
\usepackage{color,soul}

\usepackage{amsmath,amsfonts,amsthm,eucal}
\usepackage{enumerate}
\usepackage[letterpaper,margin=3cm]{geometry}
\usepackage{float}
\usepackage[title]{appendix}
\usepackage{color}
\usepackage{tabularx}
\usepackage{siunitx}
\usepackage[tableposition=below]{caption}
\usepackage{array}

\usepackage[
pdfstartview=XYZ,
bookmarks=true,
colorlinks=true,
linkcolor=blue,
urlcolor=blue,
citecolor=blue,
pdftex,
bookmarks=true,
linktocpage=true,   
hyperindex=true
]{hyperref}

\usepackage{lipsum}
\usepackage{lineno}

\usepackage[FIGBOTCAP,TABTOPCAP,bf,tight]{subfigure}

\usepackage{natbib}
\setcitestyle{authoryear}

\usepackage[pdftex]{graphicx}
\usepackage{epstopdf}
\usepackage{booktabs} 
\usepackage{tikz,mathpazo}
\usetikzlibrary{shapes.geometric, arrows}
\usepackage{caption}

\newcommand{\tensor}[1]{\ensuremath{\boldsymbol{#1}}}

\usepackage{algorithm}
\usepackage{algpseudocode}

\algdef{SE}[SUBALG]{Indent}{EndIndent}{}{\algorithmicend\ }%
\algtext*{Indent}
\algtext*{EndIndent}

\theoremstyle{remark}

\newtheorem{remark}{Remark}

\renewcommand{\vec}[1]{\ensuremath{\boldsymbol{#1}}}
\theoremstyle{definition}

\newcolumntype{M}[1]{>{\centering\arraybackslash}m{#1}}

\title{Geometric deep learning for computational mechanics Part II: Graph embedding for interpretable multiscale plasticity} 

\begin{document}

\author{Nikolaos N. Vlassis\thanks{Department of Civil Engineering and Engineering Mechanics, 
 Columbia University, 
 New York, NY 10027.     \textit{nnv2102@columbia.edu}  }       \and
WaiChing Sun\thanks{Department of Civil Engineering and Engineering Mechanics, 
 Columbia University, 
 New York, NY 10027.
  \textit{wsun@columbia.edu}   }
}

\maketitle

\begin{abstract}
The history-dependent behaviors of classical plasticity models are often driven by internal variables evolved according to phenomenological laws. 
The difficulty to interpret how these internal variables represent a history of deformation, the lack of direct measurement 
of these internal variables for calibration and validation, and the weak physical underpinning of those phenomenological
laws have long been criticized as barriers to creating realistic models. In this work, geometric machine learning on graph 
data (e.g. finite element solutions) is used as a means to establish a connection between nonlinear dimensional reduction techniques
and plasticity models. Geometric learning-based encoding on graphs allows the embedding of rich time-history data onto a low-dimensional Euclidean space
such that the evolution of plastic deformation can be predicted in the embedded feature space. 
A corresponding decoder can then convert these low-dimensional internal variables back into a weighted graph
such that the dominating topological features of plastic deformation can be observed and analyzed. 
\end{abstract}

\section{Introduction}
\label{intro}
The composition of a macroscopic plasticity model often requires the following steps. 
First, there are observations of causality relations deduced by modelers to hypothesize 
mechanisms that lead to the plastic flow. 
These causality relations along with constraints inferred 
from physics and universally accepted principles lead to mathematical equations. 
For instance, 
the family of Gurson models employs the observation of void growth to employ the yield surface \citep{gurson1977continuum}.
Crystal plasticity models relate the plastic flow with slip systems to predict the anisotropic responses of single crystals \citep{rice1971inelastic, uchic2004sample, clayton2010nonlinear, ma2020computational, ma2021atomistic}. 
Granular plasticity models 
propose theories that relate the fabric of force chains and porosity to the onset of plastic yielding and the resultant plastic flow \citep{cowin1985relationship, kuhn2015stress, wang2018multiscale, sun2022data}. 
Finally, the mathematical equations are then either used directly 
in engineering analysis and designs (e.g. the Mohr-Coulomb envelope) or are incorporated into a boundary value problem 
in which the approximation solution can be obtained from a partial differential equation solver that 
provides incremental updates of stress-strain relations.

However, a subtle but significant limitation of this paradigm is that it imposes the burdens on modelers of being able 
to describe the mechanisms \textbf{verbally} via terminologies or atomic facts (cf. \citet{daitz1953picture, griffin1964wittgenstein}. ) 
before they can convert the theoretical ideas into mathematical equations or computer algorithms.  
Our perception of mechanics is therefore limited by our language or more precisely the limited availability of descriptors 
that enable us to record hypotheses and proposed causalities \citep{Auer_2018, wang2019cooperative, wang2021non, sun2022data}. 
What if there exists a mechanism or more precisely an evolution of spatial patterns that dominate the 
macroscopic responses but there have not yet been a proper set of terminologies to describe them properly? What happens 
if the underlying mechanisms for plastic deformation or other path-dependent behaviors cannot be described in a manner 
that is \textit{simultaneously} elegant, precise, and sufficient? 

Note that macroscopic plasticity is often formulated based on the assumption that an effective medium that is homogenized 
at the macroscopic scale exists such that it yields comparable constitutive responses to the real materials. 
As such, not only the volume-averaged physical qualities, such as porosity and dislocation density but also 
how these history-dependent \textbf{patterns} evolve over space and time, may affect the overall constitutive responses. 
The evolution of \textbf{patterns}, however, is often more difficult to be precisely described in mathematical terms than 
the volume averaged or homogenized physical quantities (such as dislocation density and porosity), 
and hence less likely to be incorporated into constitutive laws directly.
The consequence is that, while many models can accurately capture the essence and general trend of the plasticity at the macroscopic scales, 
the lack of predictions on the underlying spatial patterns within the representative elementary volume makes it 
difficult to further improve the accuracy of constitutive models due to the insufficient precision afforded by the existing descriptors.  

The classical alternative to bypass this issue is to introduce additional internal variables and the corresponding phenomenological evolution laws for these 
internal variables such that, as \citet{rice1971inelastic} points out, the internal arrangement not describable through explicit state variables can be captured (cf. Section 2.4 \citep{rice1971inelastic, dafalias1976plastic}). 
For instance, \citet{karapiperis2021data} criticize the implicit and ad-hoc nature of the internal variables  
and propose the model-free paradigm as a potential alternative. This paradigm could be feasible if new data could be generated on-demand in a cost-efficient manner, or if the existing database has sufficiently dense data points to capture all types of path-dependent behaviors of an RVE. 
However, the curse of dimensionality, the lack of a proper distance structure, and the demand for a large amount of data could all be obstacles to model-free inelasticity solver, as demonstrated, for instance, in \citet{he2020physics, eggersmann2019model, carrara2020data}.
A second alternative to bypass the use of internal variables is to introduce a concurrent (e.g.\citet{fish2007concurrent, sun2014multiscale, sun2017mixed}) or hierarchical 
multiscale scheme (e.g. \citep{yvonnet2007reduced, geers2010multi, liu2016nonlocal, sun2018prediction, he2022multiscale}). 
The trade-off of these multiscale schemes is the additional cost required to upscale the constitutive responses through computational homogenization.
As such, the training of a neural network (cf. \citet{ghaboussi1991knowledge}) or other alternatives such as Gaussian processes  (e.g. \citet{fuhg2022local, frankel2022machine}) 
can provide a cost-saving way to generate surrogates that bypass the on-the-fly computational homogenization. 
Nevertheless, without a set of internal variables to represent history in a low-dimensional space, how and whether history-dependent effects are sufficiently 
captured become ambiguous. For instance, early machine learning constitutive models, which are also free of internal variables, 
may simply employ a few previous incremental strains 
as inputs to introduce a path-dependent effect \citep{ghaboussi1991knowledge, lefik2009artificial}. However, determining the optimal incremental steps involved in the predictions for a given loading rate is not trivial. \citet{wang2018multiscale} and later \citet{mozaffar2019deep} introduce long short-term memory (LSTM) networks to incorporate path dependence for constitutive laws while avoiding the vanishing gradient issues that might otherwise be encountered in the training. However, adversarial examples generated from deep reinforcement learning (cf. \citet{wang2021non}) have revealed
the difficulty of generating a robust extrapolation of predictions outside the sampling ranges. 

This paper presents a new alternative to introducing evolution laws of internal variables to solve elastoplasticity problems with 
a new focus on connecting the macroscopic internal variables with the spatial patterns of plastic deformation at the sub-scale level through machine learning. 
Instead of introducing internal variables doomed to be not explicitly describable, as stated in \citet{rice1971inelastic}, our goal here is to create a generic framework where a graph convolutional neural network may deduce 
internal variables that can double as low-dimensional descriptors 
of microstructural patterns of representative elementary volume. 
Here, we consider a multiscale modeling problem in which high-fidelity finite element simulations of microstructures or digital image correlation generate weighted graph data. 
By using the encoder of a graph-based autoencoder to embed the plastic deformation data stored in a finite element mesh onto a low-dimensional vector space, we introduce an autonomous approach to represent strain history in a low-dimensional 
Euclidean space where evolution laws can be then deduced. 
Meanwhile, the graph decoder maps this low-dimensional vector space back to the weighted graph that represents 
the plastic patterns and therefore illustrates the dominant patterns that govern the geometry of the yield surface and 
the evolution of the plastic flow \citep{demers1992non,hinton2006reducing,xu2020multi,he2021deep,bridgman2022heteroencoder}.

The rest of the paper is organized as follows. In Section~\ref{sec:graph_based_descriptors}, we discuss the representation of plastic distribution patterns as graphs,
the graph autoencoder architecture that will be utilized to generate the encoded graph-based descriptors, as well as how they will be incorporated into the neural network elastoplastic constitutive model to make 
forward interpretable predictions.
In Section~\ref{sec:descriptor_generation}, we describe the generation of the elastoplastic database of graphs for complex microstructures and the training of the graph autoencoder as well as
conduct parametric studies on the robustness of the architecture.
In Section~\ref{sec:constitutive_predictions}, we investigate the forward prediction capacity of the proposed neural network constitutive model, compare it with recurrent neural networks from the literature, and discuss the ability to decode the predicted graph-based descriptors to interpret the behavior in the microscale.
Section~\ref{sec:conclucion} provides concluding remarks.

\section{Geometric learning for graph-based multiscale plasticity descriptors}
\label{sec:graph_based_descriptors}

In this section, our objective is to explain how we generate graph-based internal variables to represent complex elastoplastic microstructures 
and how these machine learning-generated internal variables are incorporated into a component-based neural network plasticity model \citep{vlassis2022component}.
In Section~\ref{sec:graph_representation}, we discuss the interpretation of the plastic distribution of a microstructure as a node-weighted undirected graph.
In Section~\ref{sec:graph_autoencoder}, we describe the graph convolutional filter that extracts topological information from these graphs and the overall graph autoencoder architecture
for the generation of the encoded feature vector internal variables.
In Section~\ref{sec:constitutive_model}, we discuss the constitutive model components that comprise the neural network elastoplastic model.
Finally, in Section~\ref{sec:return_mapping}, we present the return mapping algorithm that will interpret these model components to make forward interpretable elastoplastic predictions. 

\subsection{Low-dimensional representation of internal history variables in a finite element mesh}
\label{sec:graph_representation}

In finite element simulations, there is no direct access to the internal variables as a smooth field. Constitutive updates 
only occur in integration points. While a projection may enable the construction of field data suitable for supervised learning
conducted via a convolution neural network, such a projection may introduce errors especially when strain 
localization occurs \citep{mota2013lie, na2019configurational, ma2022finite}. As such, we use a collection of undirected weighted graphs as an alternative to 
represent the patterns of the plastic deformation in the RVE to bypass the need of reconstructing a smooth field in the Euclidean space. 

\begin{figure}[h!]
\centering
\includegraphics[width=.75\textwidth ,angle=0]{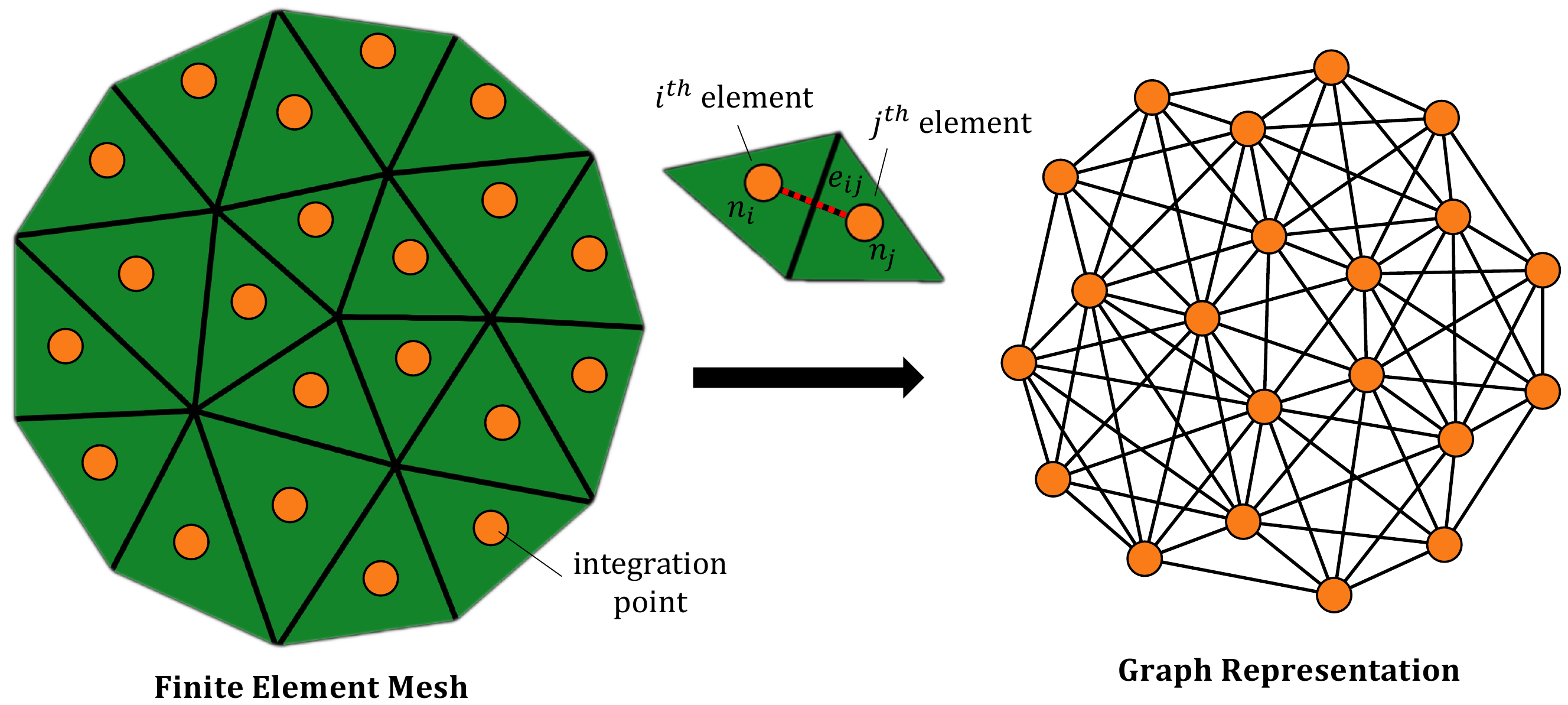} 
\caption{Interpretation of a finite element mesh as an undirected graph. The nodes of the graph represent the integration points of the mesh.
Two nodes are connected with an edge when the integration points they represent are from neighboring elements - their respective triangular elements share a side.}
\label{fig:graph_construction}
\end{figure}

Consider the body of a representative elementary volume discretized in finite elements. 
In each finite element, there is at least one integration point where the constitutive update 
is carried out. In our numerical examples, we consider
a two-dimensional triangular mesh with one integration point per element.
In this case, a pair of integration points of two adjacent finite elements can be connected by 
their shared edge (see Fig. \ref{fig:graph_construction}). 
Repeating this process connects
all integration points $\mathbb{V}$ across elements with an adjacent edge set $\mathbb{E}$ to 
form an undirected graph $\mathbb{G}=(\mathbb{V},\mathbb{E})$, a two-tuple 
that represents the topology of the integration points
where $\mathbb{V}=\left\{v_{1}, \ldots, v_{N}\right\}$ is the vertex/node set that represents the $N$ number
of integration points of the mesh
and $\mathbb{E} \subseteq \mathbb{V} \times \mathbb{V}$ is the edge set that represents the element connectivity.
The undirected unweighted graph represents the topology of the integration points and provides a data structure to store the history data to 
generate internal variables. 

We attempt to represent the irreversible/memory-dependent patterns 
stored in the vertices of the graph in a compressed low-dimensional space. 
In other words, we attempt to create a low-dimensional vector space $\mathbb{R}^{n}$ where an element $\vec{\zeta}_{i} \in \mathbb{R}^{n}$ of this space may (1) 
represent memory effects stored in these integration points in the graph $\mathbb{G}$ and (2) any vector admissible $\vec{\zeta}$ must 
fulfill the necessary conditions for the valid internal variables, which are listed below \citep{rice1971inelastic, kestin1969paradoxes}.
\begin{itemize}
\item The plastic portion of the total strain is caused by the change in internal variables for a fixed stress (and temperature). 
\item The elastic portion of the total strain is caused by the change of stress (and/or temperature) while holding the internal variable fixed. 
\end{itemize}

For a multiscale model where a surrogate model is constructed to replace the direct numerical simulations at the RVE and provides the constitutive updates, 
a feasible candidate is to consider the coordinates of the integration points and the plastic strain with the RVE as the vertex weight to form a vertex(node)-weighted graph, a 3-tuple $\mathbb{G}'(\mathbb{V}, \mathbb{E}, \vec{w})$. 
The vertex weight of a vertex-weighted graph $\vec{w}: \mathbb{V} \rightarrow \mathbb{R}^{m}$ maps an element of the vertex set onto 
a vector weight where $m=5$ for two-dimensional cases and $m=9$ for three-dimensional cases -- considering the dimensions of the position vector and the plastic strain tensor (in Voigt notation). 
We then run direct numerical simulations of the RVE to collect snapshots of this vertex-weighted graph under different loading cases and prescribed boundary conditions of the RVE, with a coverage requirement similar to those used for constructing orthogonal bases for reduced order simulations via the method of snapshots \citep{rowley2000reconstruction, 
zhong2018adaptive, zhong2021reduced}. 
Assuming that the sampling for the path-dependent behaviors is sufficient, the next step is to establish a pair of mappings between these weighted graphs and their low-dimensional representation.
This pair is obtained by training a graph autoencoder in which the encoder maps onto a vector space spanned by the collection of encoded feature vectors $\vec{\zeta}_{i}$ and the decoder maps a given vector $\vec{\zeta}$ onto a weighted graph that can, in turn, be interpreted via a finite element mesh (see Fig. \ref{fig:graph_construction}). 
As such the autoencoder is used as a tool to perform nonlinear dimensionality reduction.  
The architecture of this graph autoencoder is described in Section~\ref{sec:graph_autoencoder} .

\remark{
In this work, the plasticity graphs can be sufficiently represented with two data structures. 
The first is the adjacency matrix $\tensor{A}$, which records the connectivity of the graph. 
The adjacency matrix has dimensions $N \times N$. If nodes $n_{i}$ and $n_{j}$ are the endpoints of an edge, then the entry $A_{ij}$ is $1$.
If not, $A_{ij} = 0$. The second data structure is a node feature matrix that represents the features of every node in the graph.
The second is a node feature matrix $\tensor{X}$ that has dimensions $N \times D$ where $D$ is the number of features of a node.
Each node in the plasticity graphs has a feature vector of length $D$. For 2D simulation data, $D = 5$ while, for 3D simulation data, $D = 8$.
Each row of the node feature matrix represents the features of a node.
These two data structures will also be the actual input of the graph autoencoder described in the following section.}

\subsection{Graph autoencoder for the generation of graph-based internal variables}
\label{sec:graph_autoencoder}

The plasticity graphs described in the previous section will be the basis for the generation of the internal variables that will characterize the plastic behavior at the macroscale.
With the recent increased demand on geometric learning for graph and manifold data, many applications of autoencoders have expanded to the non-Euclidean space.
Similar to the autoencoder designed for Euclidean data, 
a graph autoencoder can perform a variety of common tasks such as, latent representation and link prediction \citep{kipf2016variational}, graph embedding \citep{pan2018adversarially}, and clustering \citep{wang2017mgae}.
This section focuses on how a graph autoencoder architecture can be used to generate internal variables that are directly connected to spatial patterns of a RVE. 
Autoencoder architectures \citep{demers1992non,hinton2006reducing,vincent2008extracting} have been used on high-dimensional data structures to perform non-linear unsupervised dimensionality reduction. 
Autoencoder architectures consist of two parts, an encoder and a decoder.
The encoder compresses the high-dimensional structure in a latent space of a much smaller dimensionality, extracting features and patterns from the structure represented as an encoded feature vector.
The decoder decompresses the encoded feature vector back to the original higher dimension attempting to accurately reconstruct the autoencoder input.
Convolutional autoencoders in the Euclidean space have been widely used in applications on images to perform - among others - 
dimensionality reduction, classification, resolution increase, and denoising tasks \citep{zeng2015coupled,lore2017llnet,chen2017deep, he2021deep}.

\begin{figure}[h!]
\centering
\includegraphics[width=.95\textwidth ,angle=0]{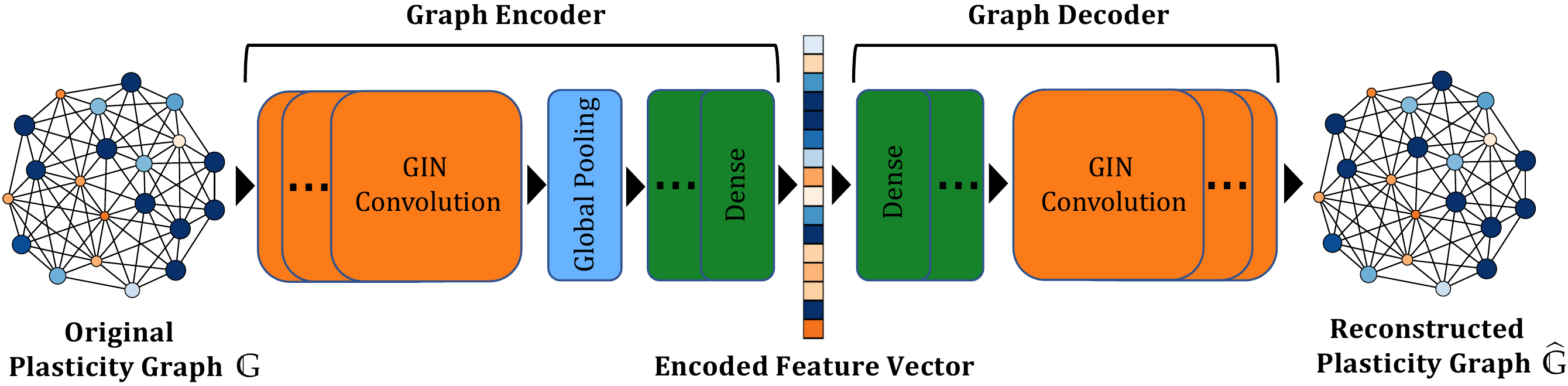} 
\caption{Graph autoencoder neural network architecture. The autoencoder encodes the original node-weighted plasticity graph into an encoded feature vector. 
The encoded feature vector is then decoded to a reconstructed plasticity graph.}
\label{fig:architecture}
\end{figure}

In this work, the plasticity simulation data are processed through a graph autoencoder $\mathcal{L}$ that aims to encode and reconstruct the plastic strain features at the graph nodes.
The autoencoder inputs a plasticity graph $\mathbb{G}$ as described in Section~\ref{sec:graph_representation} and outputs its reconstruction $\widehat{\mathbb{G}} = \mathcal{L} \left( \mathbb{G} \right)$.
The capacity of the autoencoder is two-fold and is driven by the two parts that constitute the architecture, the encoder $\mathcal{L}_{\text{enc}}$ and the decoder $\mathcal{L}_{\text{dec}}$.
The primary objective of the encoder is to represent, or encode, the graph structure so that it can be easily exploited by other machine learning models, such as the yield function and kinetic law architecture described in Section~\ref{sec:constitutive_model}.
The encoder interpolates the node feature data (topology and plastic strain tensor) that represent the entire domain of the simulation into a vector space of much smaller dimensionality.
These encoded feature vectors represent the patterns recorded during the elastoplastic loading and can be utilized as internal variables for data-driven macroscopic elasto-plastic models. 
The neural network components that constitute the constitutive update algorithm are described in Section~\ref{sec:constitutive_model}. 

The decoder decodes these encoded internal variables and reconstructs the distribution of the plastic strain within the RVE represented by the corresponding decoded weight graph. 
As such, the reconstructed graph node values is corresponding to the plastic strain values of the integration points in the finite element mesh.
This connection between internal variables and graph data makes the evolution of the internal variables interpretable through decoding. 
As shown in the latter sections, a salient feature of the proposed framework is the capacity to indirect predict 
the evolution of the high-dimensional pattern rearrangement in the microscopic scope through the kinetic law 
of the low-dimensional encoded feature vector. This treatment 
 open a new door to predict the spatial patterns of the plastic strain that are also compatible with the physical constraints of the up-scaled constitutive laws (e.g. consistency condition, incremental stress correction from plastic flow, Karush-Kuhn-Tucker conditions) upon homogenization.

\subsubsection{Graph autoencoder architecture}
\label{sec:autoencoder_architecture}

In recent years, graph neural network (GNN) techniques\citep{hamilton2017representation} based on node-neighborhood aggregation and graph-level pooling \citep{scarselli2008graph,kipf2016semi,hamilton2017inductive,defferrard2016convolutional} 
have become increasingly popular tools to embed graph data. 
In general, graph neural network layers utilize recursive aggregation or message passing methods to aggregate and interpolate the feature vectors of a node and its neighbors to compute a new feature vector.
A representation of the entire graph feature can be achieved by global pooling or other graph-level operators \citep{li2015gated,zhang2018end}.
These graph layers extract information from the graph connectivity and graph feature to be learned/encoded by other neural network layers.

The graph convolutional layers present in this work are Graph Isomorphism Network (GIN) layers introduced by \citep{xu2018powerful}.
This variant of the GNN was shown to discriminate / represent graph structures as well as the Weisfeiler-Lehman graph isomorphism test \citep{leman1968reduction}.
The Weifeiler-Lehman test checks the isomorphism between two graphs -- it tests if the graphs are topologically identical.
This is achieved by a robust injective aggregation algorithm update that maps different node neighborhoods to different unique feature vectors.
To maximize its representation capacity, a graph aggregation algorithm should be injective and not map two different neighborhoods of nodes to the same representation.
The suggested Graph Isomorphism Network borrows these aggregation concepts and generalizes the Weifeiler-Lehman test.
The GIN architecture attempts to maximize the capacity of graph neural networks by ensuring that two isomorphic graph structures are embedded in the same representation 
and two non-isomorphic ones to different representations.

The GIN layer models an injective neighborhood aggregation of multisets (features of a neighborhood of nodes) by approximating multiset functions with neural networks.
The layer's formulation is based on the assumption that a neural network can serve as a universal approximator of any function as shown by \citep{hornik1989multilayer} and, thus, can also approximate a parametrized aggregation function. The GIN layer formulation is the following:
\begin{equation}
h_{v}^{(k)}=\operatorname{MLP}^{(k)}\left(\left(1+\epsilon^{(k)}\right) \cdot h_{v}^{(k-1)}+\sum_{u \in \mathcal{N}(v)} h_{u}^{(k-1)}\right),
\label{eq:GIN_layer}
\end{equation}
where $h_{v}^{(k)}$ is the node $v$ feature representation of the $k$-th layer of the architecture, $\epsilon$ is a learnable parameter or a fixed scalar, $N$ is the number of nodes, and $\operatorname{MLP}$ is the multi-layer perceptron to learn the aggregation function approximation. The layer can also be formulated in terms of the adjacency matrix $\tensor{A}$ and the feature matrix $\tensor{X}$:
\begin{equation}
\tensor{X}^{(k)}=\operatorname{MLP}^{(k)}\left(\left(\tensor{A}+(1+\epsilon) \cdot \tensor{I}\right) \cdot \tensor{X}^{(k-1)}\right).
\label{eq:GIN_layer_matrix}
\end{equation}

This graph operator extracts features from local neighborhoods in the plasticity graph structure. 
In order to get a graph-level representation, we also utilize a graph-level pooling operation on the features of all of the nodes.
The pooling operation used was a global average pooling defined as:
\begin{equation}
\mathbf{r}^{(k)}=\frac{1}{N_{i}} \sum_{n=1}^{N_{i}} \tensor{x}_{n}^{(k-1)},
\label{eq:global_pooling}
\end{equation}
where $\mathbf{r}^{(k)}$ is the global pooled feature vector of the graph and $\tensor{x}_{n}^{(k-1)}$ is the feature vector representation of the node $n$ at the layer $k-1$.
This global feature vector will be learned and encoded by the following multi-layer perceptron architecture to create the encoded feature vector of the graph.

A schematic of the neural network architecture used on the plasticity graphs is provided in Fig.~\ref{fig:architecture}.
The neural network consists of GIN convolutional, graph pooling, and fully-connected layers.
The network's inputs include a graph adjacency matrix $\tensor{A}$ and a node feature matrix $\tensor{X}$ as described in Section~\ref{sec:graph_representation}.
The network's output is the approximated reconstruction of the node feature matrix $\widehat{\tensor{X}}$.
All the graphs in the data set have a constant connectivity matrix as the connectivity originates from the finite element mesh.
The node features of the graph correspond to the plastic strain distribution at the current time step and change during loading.

The encoder $\mathcal{L}_{\text{enc}}$ architecture compressed the high-dimensional graph into an encoded feature.
The graph adjacency matrix $\tensor{A}$ and the node feature matrix $\tensor{X}$ are input in GIN layer with 64 output channels/filters -- the dimensionality of the feature representation for every node is increased from $D$ to 64. This is achieved by setting up a fully-connected layer of 64 neurons that will serve as the $\operatorname{MLP}$ of the aggregation function as shown in Eq.~\eqref{eq:GIN_layer_matrix}.
The activation function for this layer is set to be the Rectified Linear Unit function (ReLU). The constant $\epsilon$ is set constant and equal to zero. 
The convolutional layer is followed by a global average pooling layer as described in Eq.~\eqref{eq:global_pooling} to generate a global representation of 64 features.
This is fed into a fully-connected Dense layer of 64 neurons with a ReLu activation function.
The output is connected to another Dense layer that will produce the encoded feature vector of dimension $D_\text{enc}$.

The decoder $\mathcal{L}_{\text{dec}}$ architecture decompresses the encoded feature vector back to the original graph space.
The first layer of the decoder is a Dense layer with $N \cdot D_\text{enc}$ neurons, where $N$ is the number of nodes in the graph, with a ReLU activation function.
The output of this layer is reshaped to form a $N \times D_\text{enc}$ feature matrix.
This feature matrix is passed on GIN convolutional layer along with the adjacency matrix $\tensor{A}$.
The GIN layer has a number of layers equal to $D$ -- the $\operatorname{MLP}$ approximator of the aggregation is a Dense layer with a width of $D$ neurons.
The layer has a linear activation function.
The output of the decoder is the approximated reconstructed feature matrix of the graph $\widehat{\tensor{X}}$.

The reconstruction loss function for the autoencoder is set up to minimize the discrepancy between the input node feature matrix $\tensor{X}$ 
and the output approximated feature matrix $\widehat{\tensor{X}}$ of every graph sample.
Since the coordinates of the nodes remain constant, the loss function is targeting the accuracy of the predicted plastic strain tensor at the nodes.
The loss function is modeled after a node-wise mean squared error loss function of the features.
The autoencoder function is parametrized by weights $\tensor{W}$ and biases $\tensor{b}$ such that $\mathcal{L} = \mathcal{L}\left(  \tensor{A}, \tensor{X} | \tensor{W}, \tensor{b} \right)$ 
and the training objective can be defined as:

\begin{equation}
\boldsymbol{W}^{\prime}, \boldsymbol{b}^{\prime}=\underset{\boldsymbol{W}, \boldsymbol{b}}{\operatorname{argmin}}
\left(\frac{1}{M} \sum_{k=1}^{M} \left( \frac{1}{N} \sum_{l=1}^{N} \left\|\tensor{x}^{\text{pl}}_{k,l}-\widehat{\tensor{x}}^{\text{pl}}_{k,l}\right\|_{2}^{2}\right)\right),
\label{eq:autoencoder_loss}
\end{equation}
where $M$ is the number of graph samples in the data set, $N$ is the number of graph nodes, and $\tensor{x}^{\text{pl}}_{k,l}$,$\widehat{\tensor{x}}^{\text{pl}}_{k,l}$ are 
the true and approximated plastic strain tensor components of node $l$ of graph $k$ respectively. 

The optimization of the autoencoder weights and biases is performed using the Adam optimizer \citep{kingma2014adam} with the learning rate set to $1^{-3}$ and the rest of the parameters selected as the PyTorch default ones.
The layers’ kernel weight matrices and bias vectors were initialized using the default He normal initialization \citep{he2015delving}.
The autoencoder was trained for 2000 epochs with a batch size of 20 graph samples.
The graph autoencoder architecture is built and optimized using the PyTorch neural network library \citep{NEURIPS2019_9015} and its geometric learning extension PyTorch Geometric \citep{FeyLenssen2019}.

\subsection{Supervised learning problems for the graph-enhanced constitutive model}
\label{sec:constitutive_model}

In this work, we propose the learning problems required to establish components for the constitutive laws that are integrated together to perform elastoplasticity predictions -- including the predictions of the encoded feature vector internal variables as described in Section~\ref{sec:graph_autoencoder} that will be used to interpret the spatial patterns of the RVE. 
The formulation of the Sobolev learning problem for the elasticity model component is omitted for brevity and to avoid repetitions (see
 \citet{vlassis2020geometric, vlassis2021sobolev, vlassis2022molecular} for more implementation details). A brief discussion of the training of this model for a specific RVE is provided in Section~\ref{sec:constitutive_model_training}.
The reader may also refer to \citet{vlassis2020geometric} for an implementation of a geometric learning elasticity model intended to predict anisotropic elasticity for a family of RVEs with different microstructures.

The plastic components of the neural network-based elastoplasticity framework will be driven by three neural networks, i.e., 
\begin{enumerate}
\item  a feed-forward neural network-based yield function that predicts the plastic yielding given the current stress and loading history;
\item  a recurrent neural network-based kinetic law that infers the microstructural evolution represented by the encoded feature vector from the history of macroscopic plastic strain; 
\item a feed-forward neural network that predicts the macroscopic non-associative plastic flow based on the rate of change of the encoded feature vector.
\end{enumerate}

These three components will be integrated along with the hyperelastic energy functional using a strain-space return mapping algorithm to 
predict the material's macroscopic elastoplastic behavior as described in Section~\ref{sec:return_mapping}. What follows is the formulation of the learning problems that generate these neural networks.

\subsubsection{Feed-forward neural network-based yield function}
\label{sec:yield_function}

In this work, we adopt the level set plasticity concept \citep{vlassis2021sobolev,component2021vlassis} in which a neural network is used to generate the yield function inferred from a set of point data of the homogenized stress points collected via microscale simulations.
As such, we circumvent the need to hand-craft a yield function model by replacing the construction of the yield function with a neural network level set initialization problem. To generate the yield function, 
every yield surface point cloud stress point at each time step is paired with an encoded feature vector obtained at the same instant of the RVE simulations.  
The establishment of a yield function will distinguish the elastic path-independent responses from the plastic path-dependent counterparts. 
In addition, we also impose a restriction such that 
the evolution of the encoded feature vector internal variables may only occur during the plastic loading and stop upon elastic unloading. 
Hence, it will allow for the interpreted plastic distribution in the microscale to only evolve during a plastic increment.

The yield function neural network inputs the stress state and the current accumulated plastic strain. 
Presumably, one may also establish yield function as a function of stress and encoded feature vectors where the evolution of the encoded feature vector is captured by a recurrent neural network. However, the construction of such a yield function can be complicated due to the high dimensionality. 

As such, we borrow the ideas from generalized plasticity models \citep{pastor1990generalized} where we introduce a yield function at a low-dimensional parametric space but introduce a more elaborated effort to capture the complexity of the plastic flow by predicting the path-dependent relationships among (1) the homogenized plastic strain, (2) the encoded feature vectors that represent the dominating plastic strain patterns at the RVE, and (3) the resultant plastic flow. 
As such, the yield function is only represented in the two stress invariant $p-q$ space.
Thus, we reduce the stress representation from six dimensions (symmetric stress tensor) to an equivalent two-dimensional representation $\tensor{x}(p,q)$
where $p$  is the mean pressure and $q$ the deviatoric stress invariants of the Cauchy stress tensor.

The accumulated plastic strain and the stress invariants can be seen as the coordinates of a point cloud of yielding surface $f_{\Gamma}$ samples.
In previous work, the yield stress point cloud collected from experiments would be pre-processed into a yield function level set $\phi$.
The evolution of this level set (hardening) would be predicted through a neural network that emulates the solution of the Hamilton-Jacobi level set extension problem.
The incremental solutions of this problem is the evolving level set taken at a given monotonically increasing pseudo-time $t$, which is the accumulated plastic strain internal variable $\xi$ in our case.
The yield function instance $f_n$ that corresponds to a plastic strain level $\tensor{\epsilon}^p_n$ and -- in turn an encoded feature vector $\tensor{\zeta}_n$ -- would be the level set solution of the Hamilton-Jacobi problem $\phi_n$

To generate the neural network yield function, we first recast the yield function $f$ into a signed distance function $\phi$, such that
$f(p,q,\xi) = \phi(p,q,\xi)$.
We define a neural network approximation of the level set yield function $f$ as $\widehat{f} = \widehat{f}(p,q,\xi |\tensor{W},\tensor{b})$, parametrized by weights $\tensor{W}$ and biases $\tensor{b}$.

The training objective of the neural network optimization is modeled after the minimization of the $L_2$ norm of yield surface points in the data set 
and the Eikonal equation solution that reads $|\nabla^{\tensor{x}} \phi| = 1$, while prescribing the signed distance function to 0 at $\tensor{x} \in f_{\Gamma}$. 
Minimizing the Eikonal equation solution loss term will implicitly ensure the construction of the yield function level set.
The training loss function at training samples $(\tensor{x}_i , \tensor{\zeta}_i)$ for $i\in [1,...,M]$ is defined as:
\begin{equation}
W^{\prime}, b^{\prime}=\underset{W, b}{\operatorname{argmin}}\left(\frac{1}{M} \sum_{i=1}^{M}\left(\left\|f_{i}-\widehat{f}_{i}\right\|_{2}^{2}+w \left( \left\|\nabla^{\tensor{x}} \widehat{f}_i\right\|_{2}^{2} - 1 \right)\right)\right),
\label{eq:yield_function_loss}
\end{equation}
where $w$ is a weight hyperparameter for the Eikonal equation term.
Unlike previous level set plasticity models (e.g. \citet{vlassis2021sobolev}) which focus on using Sobolev training to bypass the use of the non-associative flow rule, this modeling framework predicts 
the history dependence of the plastic flow via a neural network non-associative flow rule.
This treatment avoids the potential gradient conflict issues for multi-objective training and simplifies the neural network training \citep{bahmani2021training}. 
In this work, a feed-forward neural network is trained to predict how the current spatial patterns represented by the encoded feature vectors lead to the changes in plastic flow. Meanwhile, the encoded feature vector itself is predicted by another recurrent neural network that links the history 
of the homogenized plastic strain of a particular RVE to an encoded feature vector that represents the current plastic strain field within this RVE
 (see Section~\ref{sec:plastic_flow}). 
Directly predicting the plastic flow from phenomenological observations between the evolution of 
hand-picked microstructural descriptors and the resultant macroscopic plastic flow is not a new idea. In fact, this treatment is commonly used in generalized plasticity models
 \citep{pastor1990generalized,dafalias2004simple, wang2016identifying, sun2013unified, liu2016determining}.
The key innovation here is the replacement of the hand-picked descriptors with a generic encoded feature vector approach that can efficiently reduce the dimensions of the topological information without losing the information that might otherwise compromise the accuracy of predictions.

\subsubsection{Recurrent neural network-based kinetic law for the encoded feature vector}
\label{sec:kinetic_law}
In this work, the evolution of the encoded feature vector internal variables is characterized by a macroscopic kinetic law that takes the history of the homogenized plastic strains of an RVE as input and outputs the change of the encoded feature vector that represents the spatial patterns of plastic strain field of this specific RVE. 
We assume no prior knowledge of the exact parameterization of this kinetic law. Instead, we obtain the approximation of this kinetic law by 
training a recurrent neural network that provides the solution of a dynamical system \citep{funahashi1993approximation, bailer1998recurrent}.

Here, we hypothesize that (1) the mapping from the homogenized plastic strain and the encoded feature vector is surjective but not necessarily injective and that (2)
the evolution of the encoded feature vector solely depends on the time history of the plastic strain tensor. 
The first condition is reasonable as it is possible to have different plastic strain fields within the same RVE 
that can be homogenized to the same macroscopic plastic strain, while the second condition is adopted for simplicity. 

As such, the trained recurrent neural network takes the history of the homogenized plastic strain tensor $\tensor{\epsilon}_{\text{hist}}^{p} = \left[\tensor{\epsilon}^{p}_{n-\ell}, \ldots, \tensor{\epsilon}^{p}_{n-1}, \tensor{\epsilon}^{p}_{n}\right]$ of length $\ell$ and outputs the encoded feature vector $\tensor{\zeta}_n$ at time step $n$.
The relation between the plastic strain and the encoded feature vector is approximated by a neural network architecture defined as $\widehat{\tensor{\zeta}} = \widehat{\tensor{\zeta}} ( \tensor{\epsilon}_{\text{hist}}^{p} )$ and parametrized by weights $\tensor{W}$ and biases $\tensor{b}$. 
The training objective for samples $\tensor{\zeta}_i$ for $i \in [1, ..., M]$ is defined as:
\begin{equation}
W^{\prime}, b^{\prime}=\underset{W, b}{\operatorname{argmin}}\left(\frac{1}{M} \sum_{i=1}^{M}\left(\left\|\tensor{\zeta}_{i}-\widehat{\tensor{\zeta}}(\tensor{\epsilon}_{\text{hist},i}^{p})\right\|_{2}^{2}\right)\right), 
\end{equation}
where the back-propagation occurs through time such that the history of the plastic strain tensor is trained as time series data. 

The kinetic law neural network is utilized in the return mapping algorithm described in Section~\ref{sec:return_mapping} to predict the change of the encoded feature vector at every constitutive update.
Since the encoded feature vector is designed to change only during the plastic step, the kinetic law is only needed when the material is yielding.
The kinetic law neural network will be used in parallel with the yield function neural network described in the previous section
in the return mapping scheme to make forward predictions that are consistent with the current plastic strain. 
At every step of the elastoplastic simulation, be it in the elastic or plastic behavior, there is access to the encoded feature vector that can in turn be decoded into the original plasticity graph space to interpret the plastic strain distribution of the material. A demonstration and discussion of this capacity are described in Section~\ref{sec:graph_interpretation}.
The predicted encoded feature vector will also be utilized to predict the plastic flow direction of the current plastic step.

\subsubsection{Feed-forward neural network-based multiscale coupling plastic flow}
\label{sec:plastic_flow}

The last constitutive model component introduced in this work is a plastic flow neural network.
This network maps the encoded feature vector internal variable to the current plastic flow direction in the macroscale.
Assuming the existence of a plastic potential function $\vartheta$ and the plastic flow is non-associative, the following relationship holds, 
\begin{equation}
\dot{\tensor{\epsilon}}^{\mathrm{p}}=\dot{\lambda} \frac{\partial \vartheta}{\partial \tensor{\sigma}},
\end{equation}
where $\lambda$ is the plastic multiplier. The plastic flow direction is not derived by the yield function stress gradient to allow a more 
precise prediction of the plastic flow made possible by the newly gained knowledge of the microstructural evolution (represented by the kinetic law of the encoded feature vector). 
For the general 3D case, the stress gradient of the plastic potential can be spectrally decomposed such that:
\begin{equation}
\frac{\partial \vartheta}{\partial \tensor{\sigma}}=\sum_{A=1}^{3} g_A \tensor{n}^{A} \otimes \tensor{n}^{A}  \text{ for } A = 1,2,3,
\end{equation}
where $g_A = \frac{\partial \vartheta}{\partial \sigma_A} $ and $\sigma_A$ is the principal stress and $\tensor{n}^{A}$ the respective principal direction.

In this work, the plastic potential has not been explicitly reconstructed from the plastic flow, although it might be possible as demonstrated in \cite{vlassis2022molecular}.  Instead, a generalized plasticity framework is adopted. 
As such, the mapping from the change of the encoded feature vector $\tensor{\delta \zeta}$ to the plastic flow $\tensor{g}$ is, approximated by a neural network architecture defined as $\widehat{\tensor{g}} = \widehat{\tensor{g}} ( \tensor{\delta \zeta} )$ and parametrized by weights $\tensor{W}$ and biases $\tensor{b}$. 
The training objective for samples $\tensor{g}_i$ for $i \in [1, ..., M]$ is defined as:
\begin{equation}
W^{\prime}, b^{\prime}=\underset{W, b}{\operatorname{argmin}}\left(\frac{1}{3M} \sum_{i=1}^{M} \sum_{A=1}^{3} \left(\left\|g_{A,i}-\widehat{g}_{A,i}(\tensor{\delta \zeta}_{i})\right\|_{2}^{2}\right)\right), 
\end{equation}
where $\tensor{\delta \zeta}_{i}$ is the interpolated change of the encoded feature vectors, which can be approximated via the Backward or Forward Euler method. 
Note that, 
in contrast to the kinetic law that is approximated via a recurrent neural network (see Section~\ref{sec:kinetic_law}), we assume that the mapping from the encoded feature vector to the macroscopic plastic flow can be sufficiently represented with a feed-forward neural network. 
This is a reasonable because the encoded feature vector is used to represent the entire plastic strain field within an RVE and 
the plastic flow can simply be inferred by homogenizing the graph data decoded from the encoded feature vector.
Presumably, one may even bypass the training of neural network plastic flow predictor by performing exactly this calculation. 
This learning problem is nevertheless introduced for ease of implementation. As demonstrated in Section~\ref{sec:constitutive_model_training}, 
training a sufficiently accurate neural network predictor for plastic flow is feasible.

\subsection{Return mapping algorithm}
\label{sec:return_mapping}

In this section, we provide the implementation details for the return mapping algorithm to make forward elasoplasticity predictions.
The fully implicit stress integration algorithm allows for the incorporation of the graph-based internal variables generated from the autoencoder architecture as described in Section~\ref{sec:graph_autoencoder} in the prediction scheme.
The return mapping algorithm is designed in the principal strain space is described in Algorithm~\ref{algo:return_map_algorithm}.

This formulation of the return mapping algorithm requires all the strain and stress measures are in principal axes.
However, this is not limiting for the choice of strain and stress space formulation of the constitutive law components as the framework allows for coordinate system transformation through 
automatic differentiation. The automatic differentiation is facilitated with the use of the Autograd library \citep{maclaurin2015autograd}. 
The yield function is written in a two stress invariant formulation -- $(p,q)$ as described in Section~\ref{sec:yield_function}.
Through automatic differentiation and a series of chain rules the constitutive model predictions are expressed in the principal axes, allowing for any invariant formulation of the yield function during training.

The elastoplastic behavior is modeled through a predictor-corrector scheme that integrates the elastic prediction with the corrections by the yield function neural network.
It is noted that the elastic update predictions for the hyperelastic energy functional and the plasticity terms encountered in the return mapping algorithm are evaluated as neural network predictions using the offline trained energy functional, yield function, and kinetic law using the Tensorflow \citep{abadi2016tensorflow} and Keras \citep{chollet2015keras} machine learning libraries. 
The hyperelastic neural network is based on the prediction of an energy functional with interpretable first-order and second-order derivatives (stress and stiffness respectively).
The plasticity neural networks utilized in this algorithm are described in Section~\ref{sec:constitutive_model}. 
Besides the capacity to predict the values of the approximated functions, these libraries also allow for the automatic evaluation of the approximated function derivatives that are required to perform the return mapping constitutive updates and constructing the local Newton-Raphson tangent matrix as well as the necessary coordinate set transformation chain rules. 

\begin{algorithm}
	\caption{Return mapping algorithm in strain-space for encoded feature vector internal variable plasticity.}\label{algo:return_map_algorithm}
\begin{algorithmic}
\renewcommand{\thealgorithm}{}
		\Require Hyperelastic energy functional $\widehat{\psi}^{\mathrm{e}}$ neural network, yield function $\widehat{f}$ neural network, the encoded feature vector neural network $\widehat{\tensor{\zeta}}$, and the plastic flow network $\widehat{\tensor{g}}$.
		\State \textbf{1.} Compute trial elastic strain
	    \Indent
		\State Compute $\tensor{\epsilon}_{n+1}^{\mathrm{e \, tr}}=\tensor{\epsilon}_{n}^{\mathrm{e}}+\Delta \tensor{\epsilon}$.
		\State Spectrally decompose $\tensor{\epsilon}_{n+1}^{\mathrm{e \, tr}}=\sum_{A=1}^{3} \epsilon_{A}^{\mathrm{e \, tr}} \tensor{n}^{\operatorname{tr},A} \otimes \tensor{n}^{\operatorname{tr},A}$.
		\EndIndent
		
		\State \textbf{2.} Compute trial elastic stress
		\Indent
		\State Compute $\sigma_{A}^{\mathrm{tr}}=\partial \widehat{\psi}^{\mathrm{e}} / \partial \epsilon_{A}^{\mathrm{e}}$ for $A = 1,2,3$ and the corresponding $p^{\operatorname{tr}}, q^{\operatorname{tr}}$ at  $\epsilon_{n+1}^{\mathrm{e \, tr}}$.
		\EndIndent
		
		\State \textbf{3.} Check yield condition and perform return mapping if loading is plastic
		\Indent
		\If{$\widehat{f}\left(p^{\operatorname{tr}}, q^{\operatorname{tr}}, \xi_{n}\right) \leq 0 $}
			\State Set $\tensor{\sigma}_{n+1}=\sum_{A=1}^{3} \sigma_{A}^{\operatorname{tr}} \tensor{n}^{\operatorname{tr},A} \otimes \tensor{n}^{\operatorname{tr},A}$ and exit.
		\Else
			\State Compute encoded feature vector $\tensor{\zeta}_{n} = \widehat{\tensor{\zeta}}(\tensor{\epsilon}^p_{\text{hist},n})$. 
			\State Compute plastic flow direction $\frac{\partial \widehat{\vartheta}}{\partial \sigma_{A}}=\widehat{\tensor{g}}(\tensor{\delta \zeta}_{n})$ for $A = 1,2,3$.
			\State Solve for $\epsilon_{1}^{\mathrm{e}}, \epsilon_{2}^{\mathrm{e}}, \epsilon_{3}^{\mathrm{e}}$, and $\xi_{n+1}$ such that $\widehat{f}\left(p,q, \xi_{n+1}\right) = 0 $.
			\State Compute $ \tensor{\sigma}_{n+1}=\sum_{A=1}^{3}\left(\partial \widehat{\psi}^{\mathrm{e}} / \partial \epsilon_{A}^{\mathrm{e}}\right) \tensor{n}^{\operatorname{tr},A} \otimes \tensor{n}^{\operatorname{tr},A}$ and exit.
		\EndIf	
		\EndIndent
\end{algorithmic}
\end{algorithm}

The return mapping also incorporates a non-associative flow rule to update the plastic flow direction instead of using the stress gradient of the yield function.
This is achieved by incorporating the predictions of the plastic flow $\widehat{\tensor{g}}$ network described in Section~\ref{sec:plastic_flow}.
Through the local iteration scheme, the solution for the true elastic strain values can be retrieved using the solved for discrete plastic multiplier $\Delta \lambda$ 
and the predicted flow direction as follows:
\begin{equation}
\epsilon_{A}^{\mathrm{e}}=\epsilon_{A}^{\mathrm{e} \operatorname{tr}}-\Delta \lambda \frac{\partial \widehat{\vartheta}}{\partial \sigma_{A}}=\epsilon_{A}^{\mathrm{e} \operatorname{tr}}-\Delta \lambda \widehat{g}_A(\widehat{\tensor{\delta \zeta}}), \quad A=1,2,3 .
\label{eq:elastic_strain_update}
\end{equation}

The return mapping algorithm requires a hyperelastic energy functional neural network $\widehat{\psi}^{\mathrm{e}}$, 
a yield function $\widehat{f}$, a kinetic law $\widehat{\tensor{\zeta}}$, and a plastic flow $\widehat{\tensor{g}}$ 
neural network that are pre-trained offline.
Given the elastic strain tensor at the current loading step, a trial elastic stress state is calculated using the hyperelastic energy functional neural network.
The yield condition is checked for the trial elastic stress state and the current plastic strain level.
If the predicted yield function is positive, the trial stress is the in the elastic region and
is the actual stress. The encoded feature vector remains constant.
If the yield function is non-positve the trial stress is in the inadmissible stress region
and an Newton-Raphson optimization scheme is utilized to correct the stress prediction.
The current encoded feature vector is predicted from the time history of plastic strain tensors and is used to predict the current plastic flow directions.
The goal of the return mapping algorithm is to solve for the prinicipal elastic strains and the plastic strain 
such that the predicted yield function is equal to zero and the stress updates is consistent with the plastic flow. 
The encoded feature vector at every step can be converted back into the corresponding weighted graph via 
 the graph decoder $\mathcal{L}_{\text{dec}}$ neural network. This weighted graph can be converted back into information 
 in a finite element mesh and therefore enable us to interpret the microstructure.
  
It is noted that the return mapping algorithm is formulated via the principle direction is provided in this section for the generalization purpose.
This setting is sufficient for isotropic materials. In our numerical examples, we only introduce two-dimensional cases to 
illustrate the ideas for simplicity. The generalization of the return mapping algorithm for anisotropic materials is straightforward, 
but the training of the yield function and the plastic flow model in the higher dimensional parametric space is not trivial. 
This improvement will be considered in the future but is out of the scope of this study.  

\section{Training of interpretable graph embedding internal variables} 
\label{sec:descriptor_generation}

In this section, we demonstrate the procedure of embedding field simulation data to construct graph-based internal variables.
A graph convolutional autoencoder is used to compress the graph structures that carry the plastic deformation distribution of a microstructure.
In Section~\ref{sec:data_generation}, we demonstrate the process of generating the plasticity data through finite element method (FEM) simulations and post-processing them into weighted graph structures.
In Section~\ref{sec:autoencoder_training}, we showcase the performance capacity of the autoencoder architecture as well as its ability to reproduce the plasticity graph structures.
Finally, in Section~\ref{sec:mesh_sensitivity}, we perform a sensitivity training test for the autoencoder architecture on different FEM meshes for the same microstructure.

\subsection{Generation of the plasticity graph database}
\label{sec:data_generation}

\begin{figure}[h!]
\centering
\includegraphics[width=.75\textwidth ,angle=0]{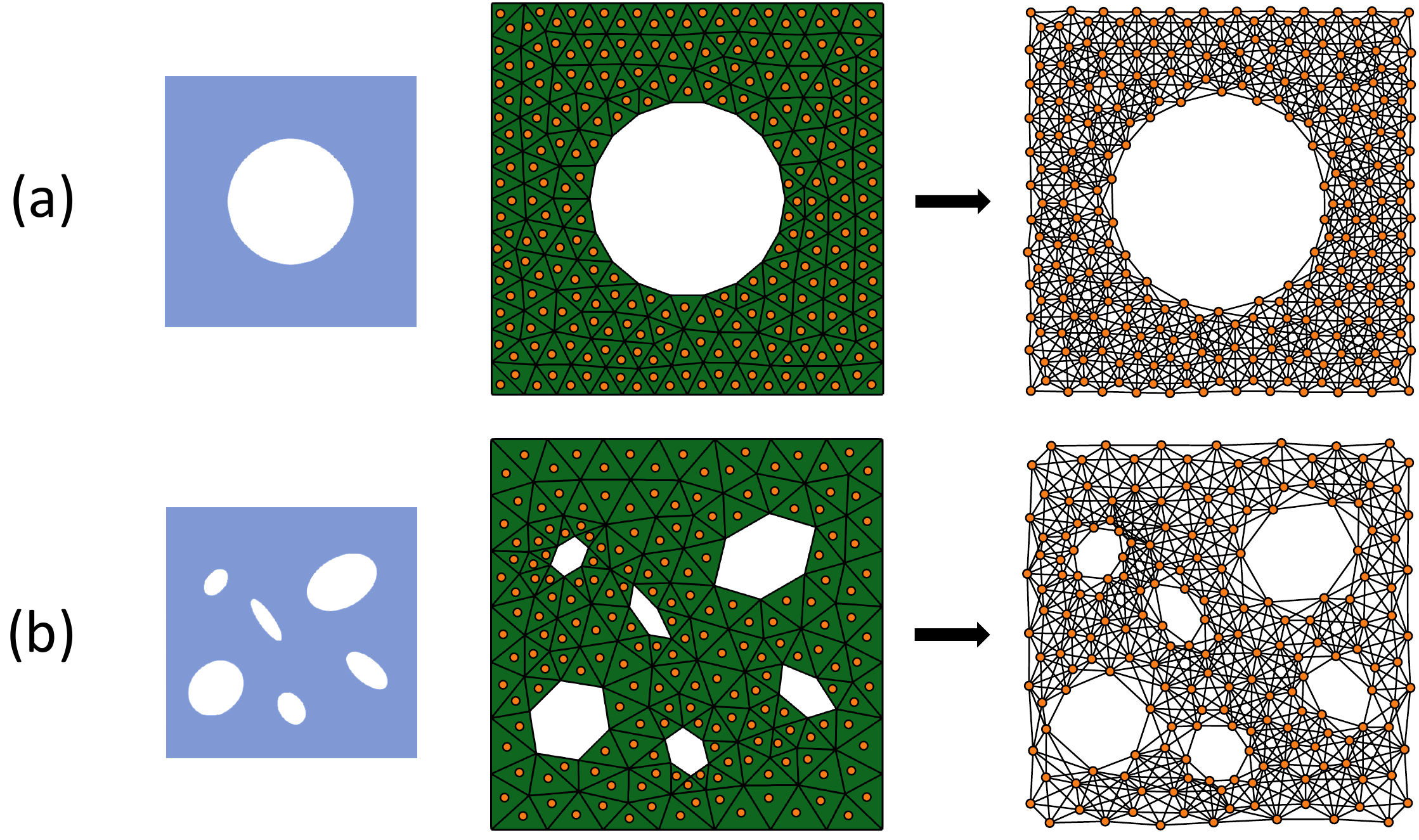} 
\caption{Two microstructures represented as a finite element mesh and an equivalent node-weighted undirected graph as described in Section~\ref{sec:graph_representation}.}
\label{fig:mesh_A_B}
\end{figure}

In this work, the autoencoders used for the generation of the graph-based internal variables and the neural network constitutive models used for the forward predictions 
are trained on data sets generated by FEM elastoplasticity simulations.
To test the autoencoders' capacity to generate encoded feature vectors regardless of the microstructure and plastic strain distribution patterns the FEM mesh represents,
we test the algorithm with two microstructures of different levels of complexity.
The two microstructures A and B are demonstrated in Fig.~\ref{fig:mesh_A_B} (a) and (b) respectively.
The outline of the microstructures is a square with a side of $1 \, \text{m}$.
This figure also shows the meshing of the two microstructures.
The microstructures A and B are discretized by 250 and 186 triangular elements respectively with one integration point each.
An investigation of different mesh sizes and the sensitivity of the encoded feature generation is demonstrated in Section~\ref{sec:mesh_sensitivity}.
Each integration point of mesh corresponds to a node in the equivalent graph (also shown in Fig.~\ref{fig:mesh_A_B}).
The integration points of the neighboring elements - elements that share an edge - are connected with an edge in the constructed graph as described in Section~\ref{sec:graph_representation}.

The constitutive model selected for the local behavior at the material points was linear elasticity and J2 plasticity with isotropic hardening.
The local behavior is predicted with an energy minimization algorithm that is introduced in \cite{miehe2002strain}.
The local optimization algorithm is omitted for brevity.
The local linear elastic material has a Young's modulus of $E = 2.0799 \text{MPa}$ and a Poisson ratio of $\nu = 0.3$. 
The local J2 plasticity has an initial yield stress of $100 \text{kPa}$ and a hardening modulus of $H = 0.1 E$.
During the simulation, the elastic, plastic, and total strain as well as the hyperelastic energy functional and stress are saved for every integration point.
The recorded plastic strain tensor along with the initial integration points coordinates will be used as the node weight vector for the plasticity graphs as described in Section~\ref{sec:graph_representation}. 
The strain, energy, and stress values will be volume averaged and used to train the neural network constitutive models for the homogenized response as described in Section~\ref{sec:constitutive_model}.

To capture varying patterns of distribution of plastic strain, the finite element simulations were performed under various combinations of uniaxial and shear loading.
The loading was enforced with displacement boundary conditions applied to all the sides of the mesh for both microstructures A and B. 
The combinations of displacement boundary conditions are sampled by rotating a loading displacement vector from $0^{\circ}$ to $90^{\circ}$ whose components are the uniaxial displacements in the two directions for the pure axial displacement cases and the uniaxial and shear displacements for the combined
uniaxial and shear loading.
The maximum displacement magnitude for axial and shear loading vector components are $u_{\text{goal}} = 1.5 \times 10^{-3} \, \text{meters}$.
We sample a total of 100 loading combinations/FEM simulations for each microstructure.
During each of these simulations, we record the constitutive response at every material point and post-process it as a node-weighted graph and a volume average response.
For every simulation, we record 100 time steps, thus collecting 10000 training sample pairs of graphs and homogenized responses for each microstructure.

\subsection{Training of the graph autoencoder}
\label{sec:autoencoder_training}

\begin{figure}[h!]
\centering
\includegraphics[width=.70\textwidth ,angle=0]{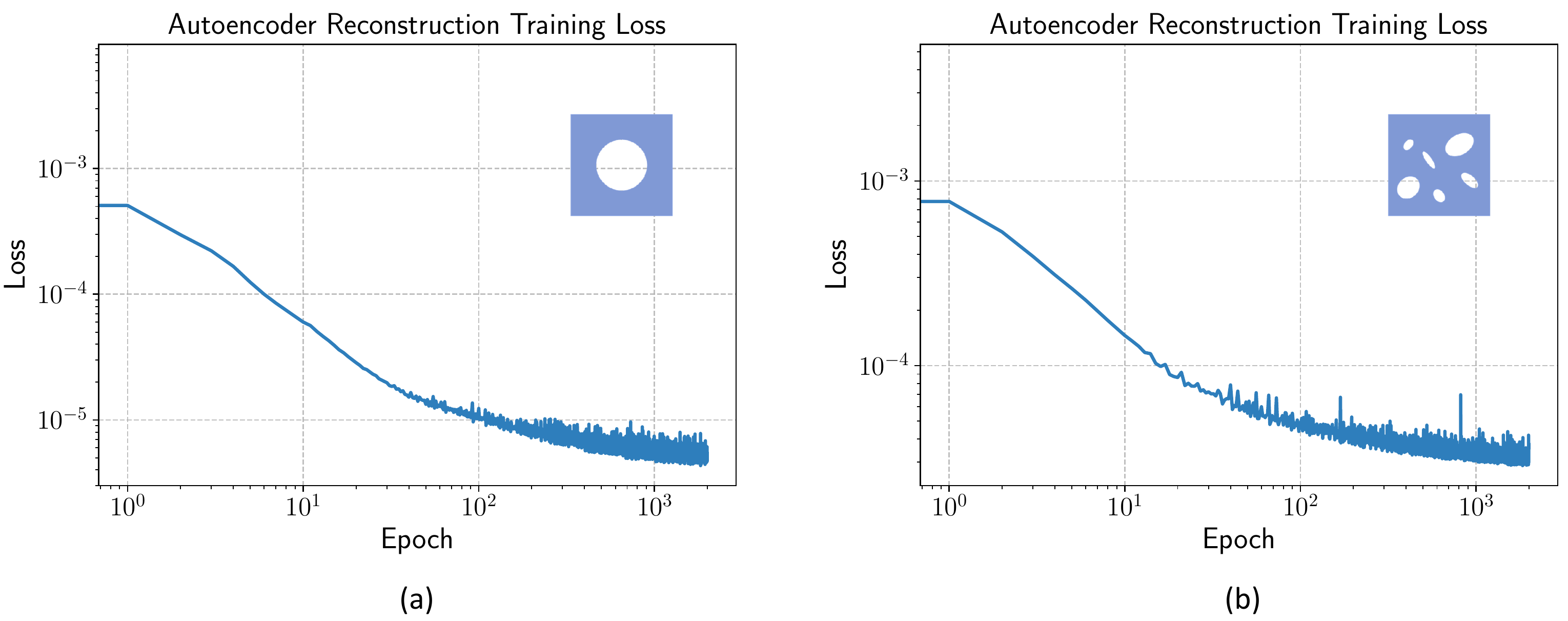} 

\caption{Autoencoder reconstruction training loss for microstructures A and B (a and b respectively) as defined in Eq.~\ref{eq:autoencoder_loss}.}
\label{fig:autoencoder_loss}
\end{figure}

In this section, we demonstrate the training performance of the autoencoder architecture on the two microstructure data sets described in Section~\ref{sec:data_generation}.
We also show the capacity of the autoencoder to reproduce the plasticity graphs in the training samples.
The autoencoder layer architecture and training procedure for both data sets are described in Section~\ref{sec:autoencoder_architecture}.
The dimension of the encoded feature vector in this example is set to $D = 16$. An examination of the effect of the encoded feature vector size is described in Section~\ref{sec:mesh_sensitivity}.

\begin{figure}[h!]
\centering
\includegraphics[width=.85\textwidth ,angle=0]{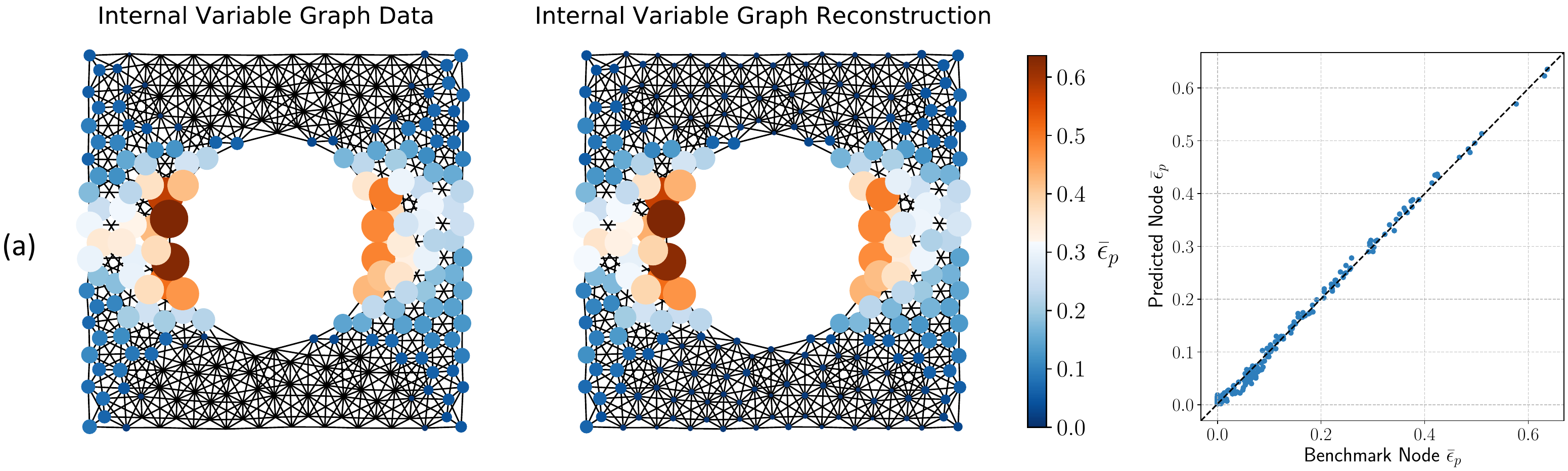} 
\includegraphics[width=.85\textwidth ,angle=0]{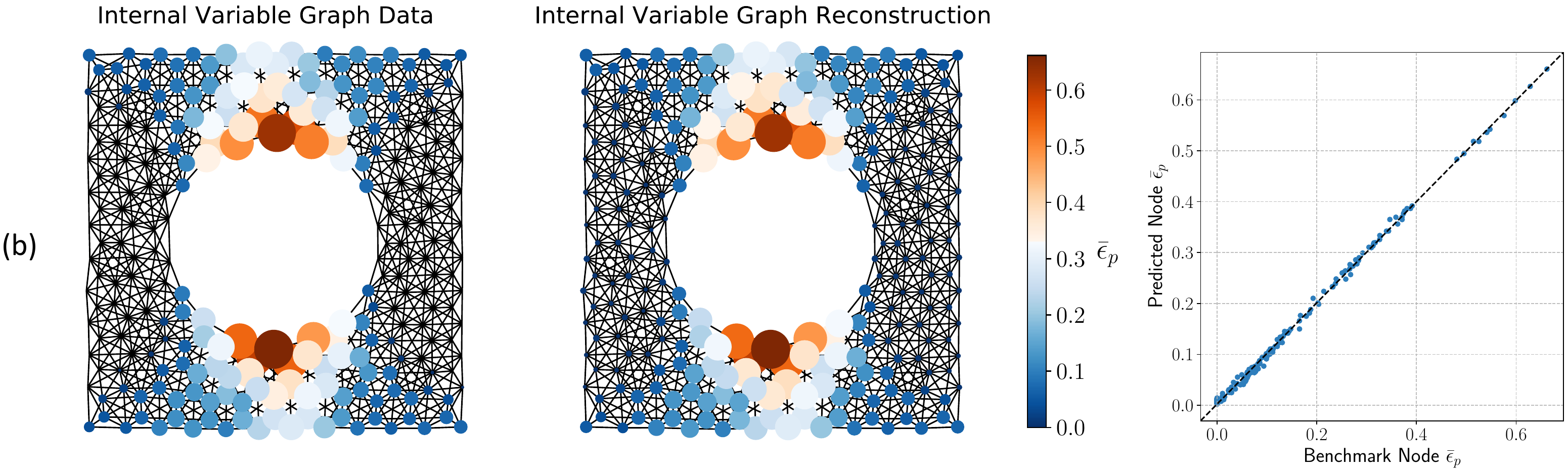} 

\caption{Prediction of the autoencoder architecture of the microstructure A for two loading paths (a and b). The graph node size and color represent the magnitude of the accumulated plastic strain $\overline{\epsilon}_p$. The node-wise predictions for $\overline{\epsilon}_p$ are also demonstrated. }
\label{fig:autoencoder_prediction_graphA}
\end{figure}

\begin{figure}[h!]
\centering
\includegraphics[width=.85\textwidth ,angle=0]{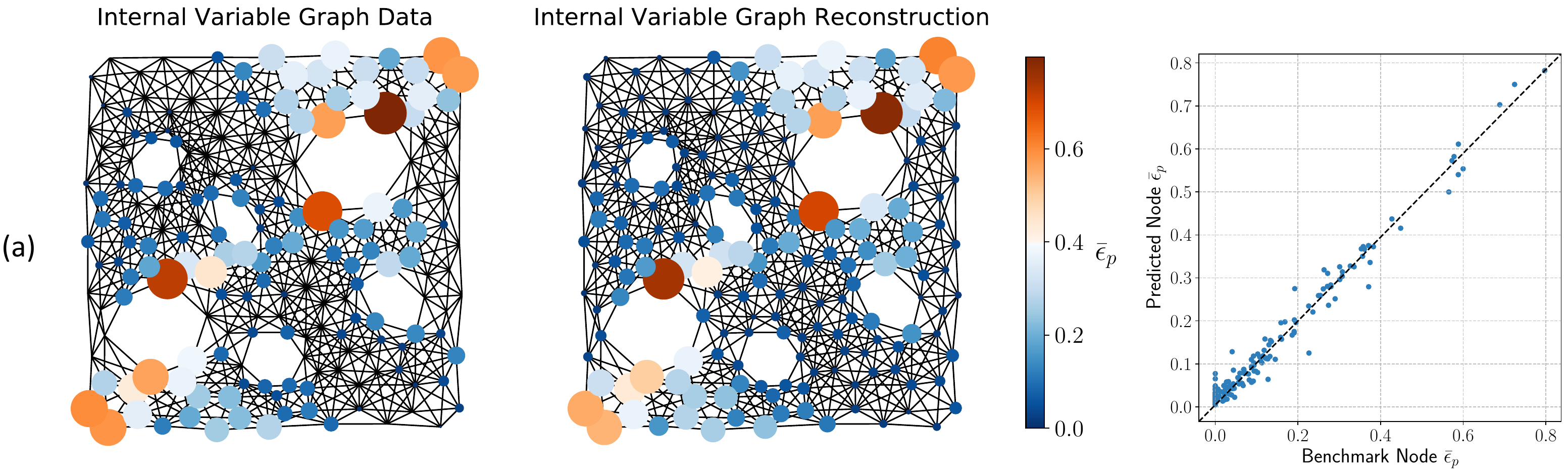} 
\includegraphics[width=.85\textwidth ,angle=0]{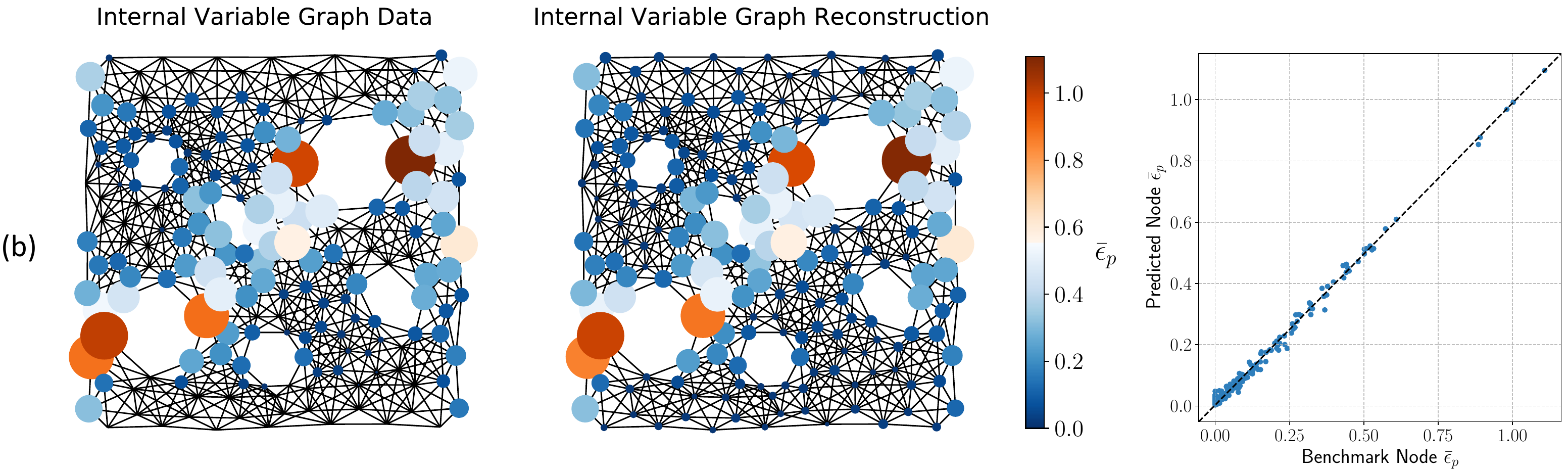} 

\caption{Prediction of the autoencoder architecture of the microstructure B for two loading paths (a and b). The graph node size and color represent the magnitude of the accumulated plastic strain $\overline{\epsilon}_p$. The node-wise predictions for $\overline{\epsilon}_p$ are also demonstrated. }
\label{fig:autoencoder_prediction_graphB}
\end{figure}

The training curves for the autoencoder's reconstruction loss function in Eq.\eqref{eq:autoencoder_loss} is shown in Fig.~\ref{fig:autoencoder_loss}.
The autoencoder appears to have similar loss function performance for both microstructures.
The autoencoder performs slightly better for microstructure A as it is tasked to learn and reproduce patterns for a seemingly simpler microstructure compared to 
microstructure B. The training loss curves in this figure demonstrate the overall performance of the autoencoder architecture -- the encoder and decoder components of the architecture
are trained simultaneously.
The capacity of the autoencoder to reconstruct the plasticity distribution patterns is explored in Fig.~\ref{fig:autoencoder_prediction_graphA} and Fig.~\ref{fig:autoencoder_prediction_graphB} 
for microstructures A and B respectively.
In these figures, we showcase the reconstruction capacity of the plastic strain for two different time steps for each microstructure.
The time steps selected are from two different loading path combinations resulting in different plasticity graph patterns.
We demonstrate how the autoencoder can reproduce these patterns by comparing the internal variable graph data -- the autoencoder input -- with the graph reconstruction -- the autoencoder output.
We also show the accuracy of the node-wise prediction of the accumulated plastic strain for these microstructures.
It is noted that the autoencoder predicts the values of the full plastic strain tensor at the nodes.
However, these plots show the accumulated plastic strain $\overline{\epsilon}_p$ values calculated from the predicted strain tensor at the nodes for easier visualization.

\begin{figure}[h!]
\centering
\includegraphics[width=.7\textwidth ,angle=0]{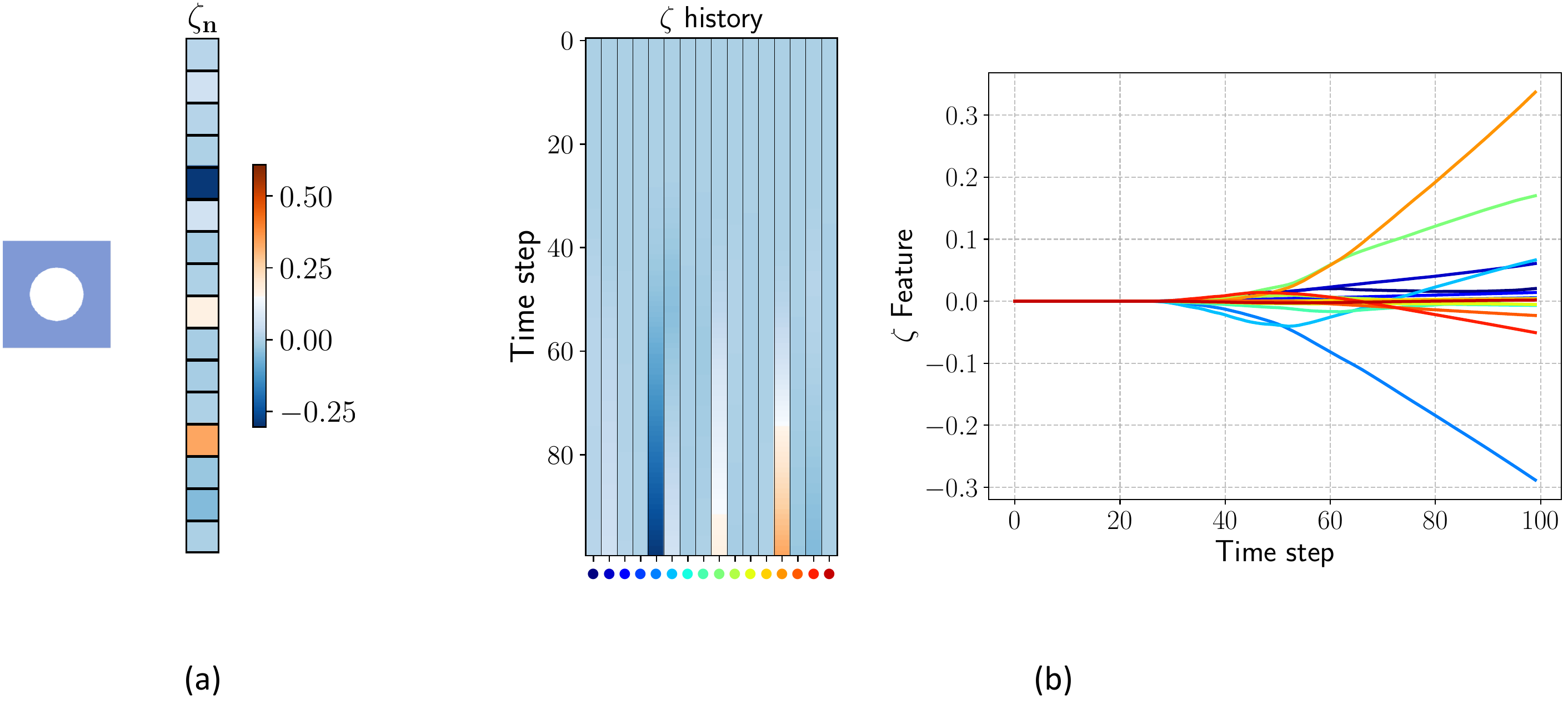} 
\includegraphics[width=.7\textwidth ,angle=0]{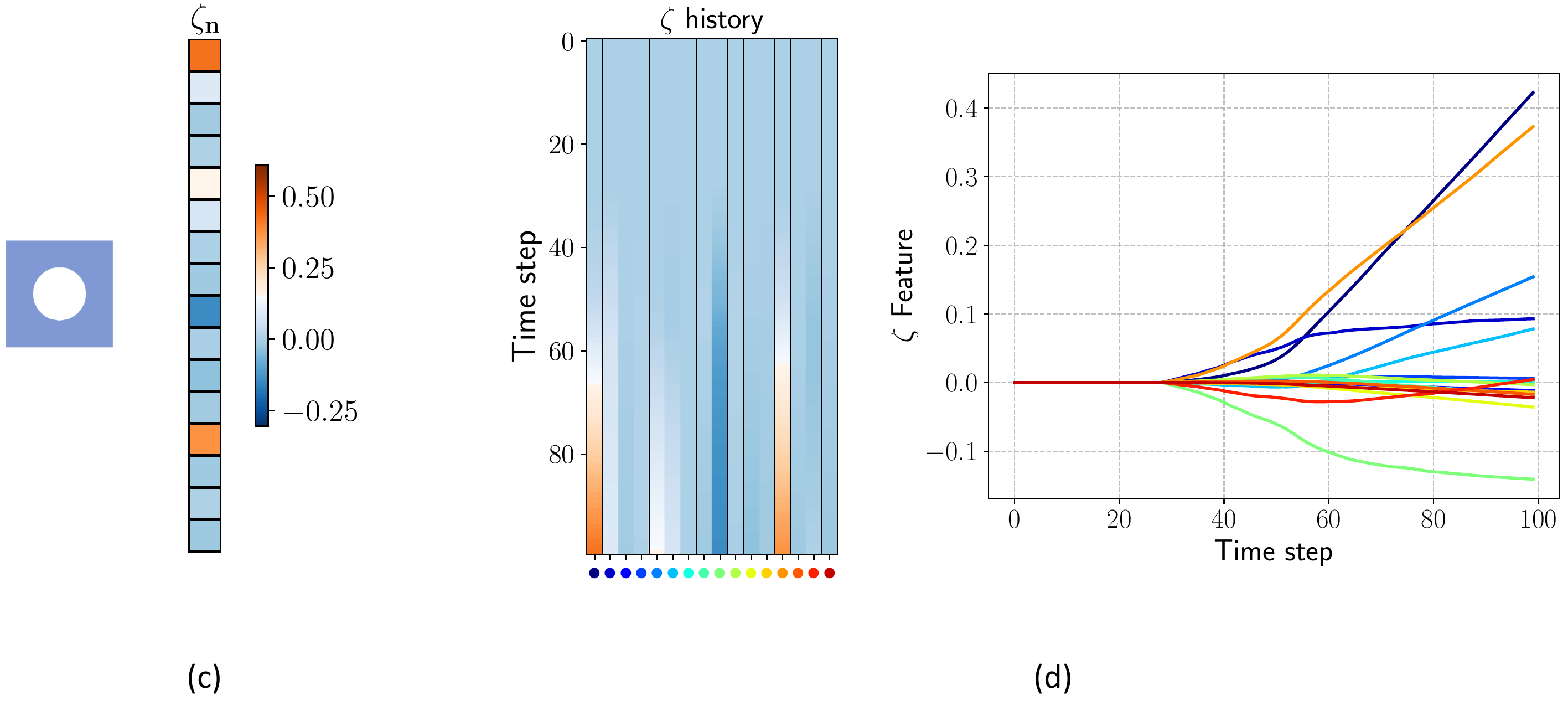} 

\caption{Prediction of encoded feature vector $\tensor{\zeta}$ by the encoder $\mathcal{L}_{\text{enc}}$ for microstructure A. 
(a,c) The encoded feature vector $\tensor{\zeta}_n$ for a single time step for the plastic graphs shown in Fig.~\ref{fig:autoencoder_prediction_graphA}a and Fig.~\ref{fig:autoencoder_prediction_graphA}b respectively.
(b,d) The encoded feature vector $\tensor{\zeta}$ history for all the time steps in the loading paths of Fig.~\ref{fig:autoencoder_prediction_graphA}a and Fig.~\ref{fig:autoencoder_prediction_graphA}b respectively.}
\label{fig:graphA_codes}
\end{figure}

\begin{figure}[h!]
\centering
\includegraphics[width=.7\textwidth ,angle=0]{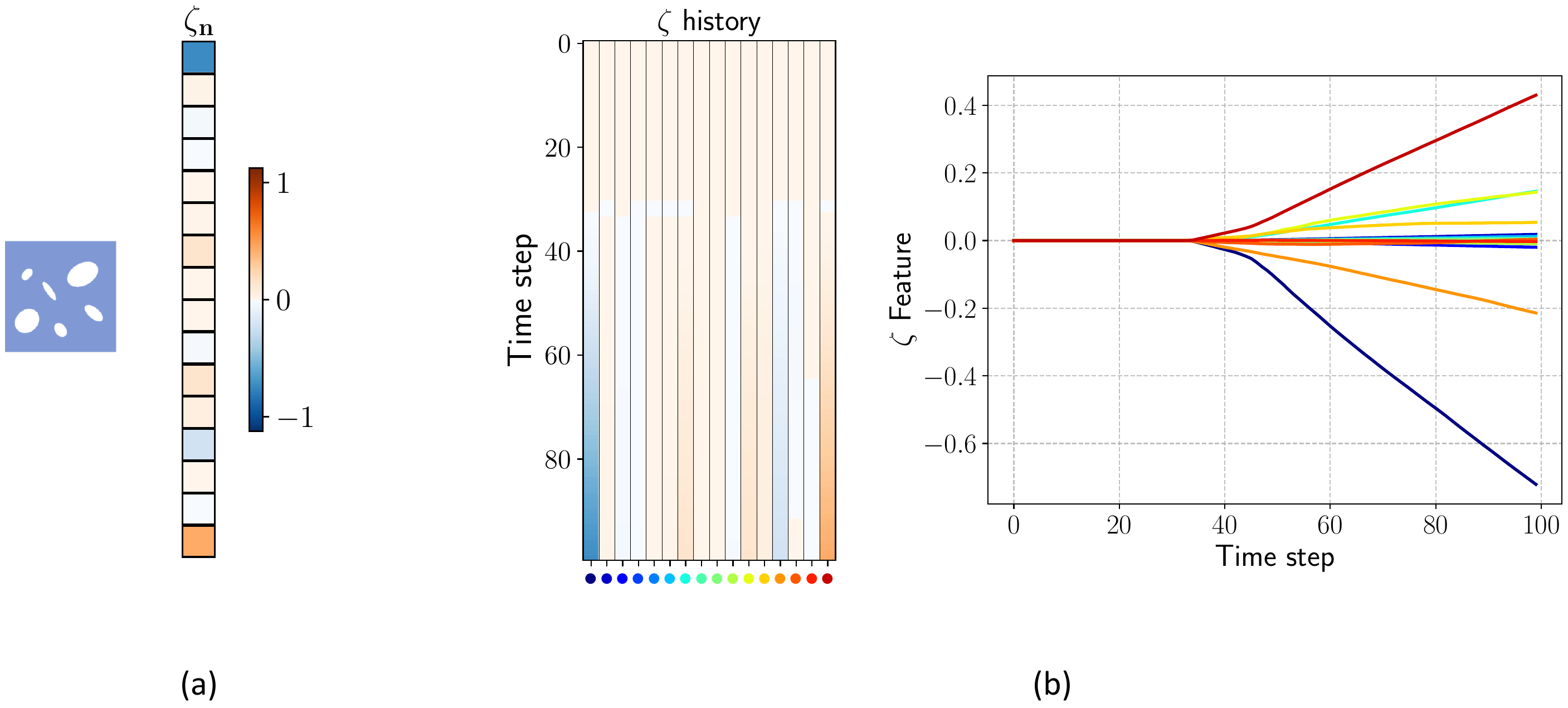} 
\includegraphics[width=.7\textwidth ,angle=0]{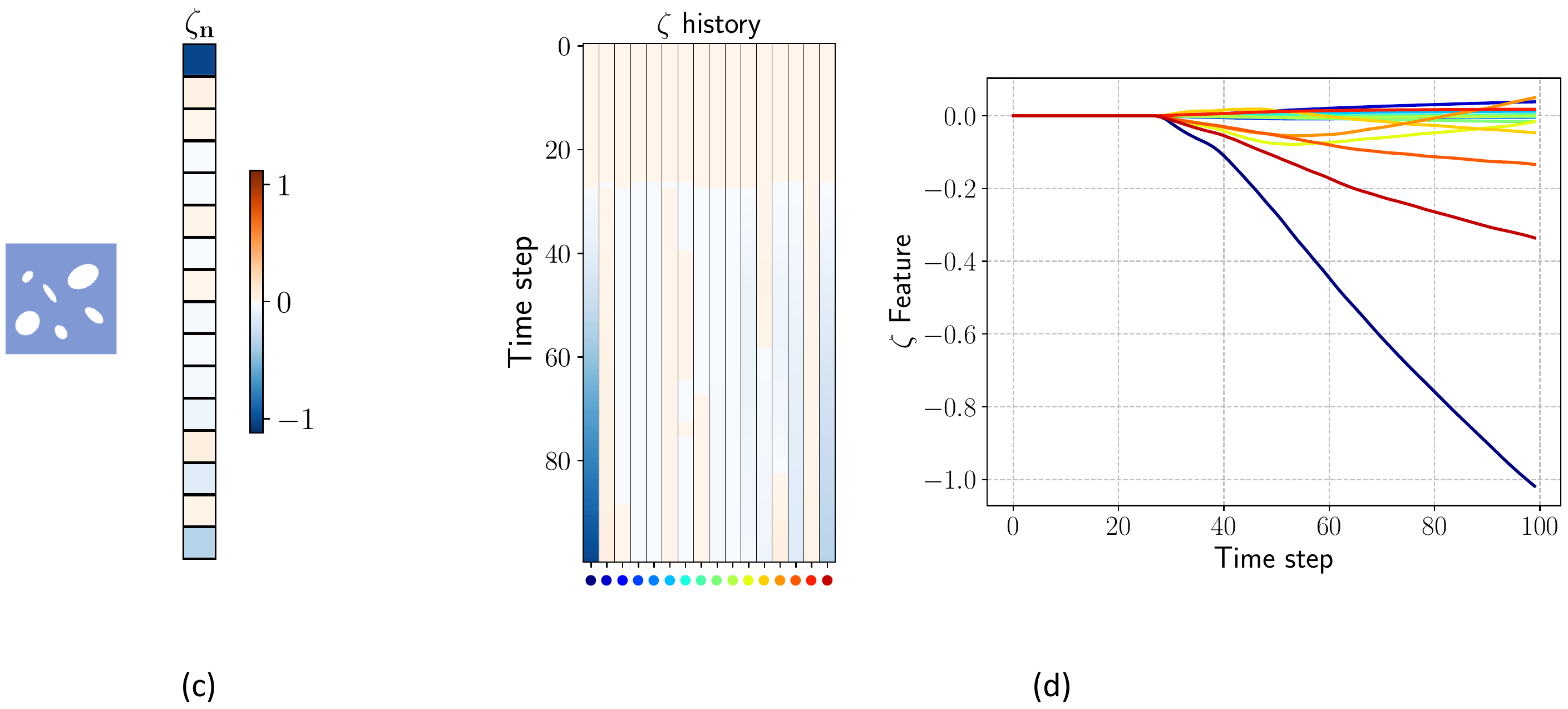} 
\caption{Prediction of encoded feature vector $\tensor{\zeta}$ by the encoder $\mathcal{L}_{\text{enc}}$ for microstructure B. 
(a,c) The encoded feature vector $\tensor{\zeta}_n$ for a single time step for the plastic graphs shown in Fig.~\ref{fig:autoencoder_prediction_graphB}a and Fig.~\ref{fig:autoencoder_prediction_graphB}) respectively.
(b,d) The encoded feature vector $\tensor{\zeta}$ history for all the time steps in the loading paths of Fig.~\ref{fig:autoencoder_prediction_graphB}a and Fig.~\ref{fig:autoencoder_prediction_graphB}b respectively.}
\label{fig:graphB_codes}
\end{figure}

The autoencoder architecture provides the flexibility of utilizing its two components, the encoder $\mathcal{L}_{\text{enc}}$ and the decoder $\mathcal{L}_{\text{dec}}$, separately.
In this section, we demonstrate the encoder's ability to process the high-dimensional graph structure in encoded feature vector $\tensor{\zeta}$ time histories.
In Fig.~\ref{fig:graphA_codes} (a) $\&$ (c) and Fig.~\ref{fig:graphB_codes} (a) $\&$ (c), we show the predicted encoded feature vector $\tensor{\zeta}_n$ for the $n$-th time step of a loading path for microstructures A and B respectively.
These encoded feature vectors specifically correspond to the graphs shown in Fig.~\ref{fig:autoencoder_prediction_graphA} and Fig.~\ref{fig:autoencoder_prediction_graphB} respectively.
In Fig.~\ref{fig:graphA_codes} (b) $\&$ (d) and Fig.~\ref{fig:graphB_codes} (b) $\&$ (d), we demonstrate the time series of plasticity graphs encoded in time series of encoded feature vectors.
It is highlighted that the encoded feature vector values do not change during the elastic/path-independent part of the loading.
This is directly attributed to the fact that the plasticity graph is constant (zero plastic strain at the nodes) before yielding for all the time steps and will be further discussed in Section~\ref{sec:constitutive_predictions}.
The benefit of separately using the decoder $\mathcal{L}_{\text{dec}}$ as a post-processing step to interpret the predicted encoded feature vectors is also explored in the following sections.

\subsection{Mesh sensitivity and Encoded Feature Vector dimension}
\label{sec:mesh_sensitivity}

\begin{figure}
\centering
\begin{minipage}{.5\textwidth}
  \centering
  \includegraphics[width=1.2\linewidth]{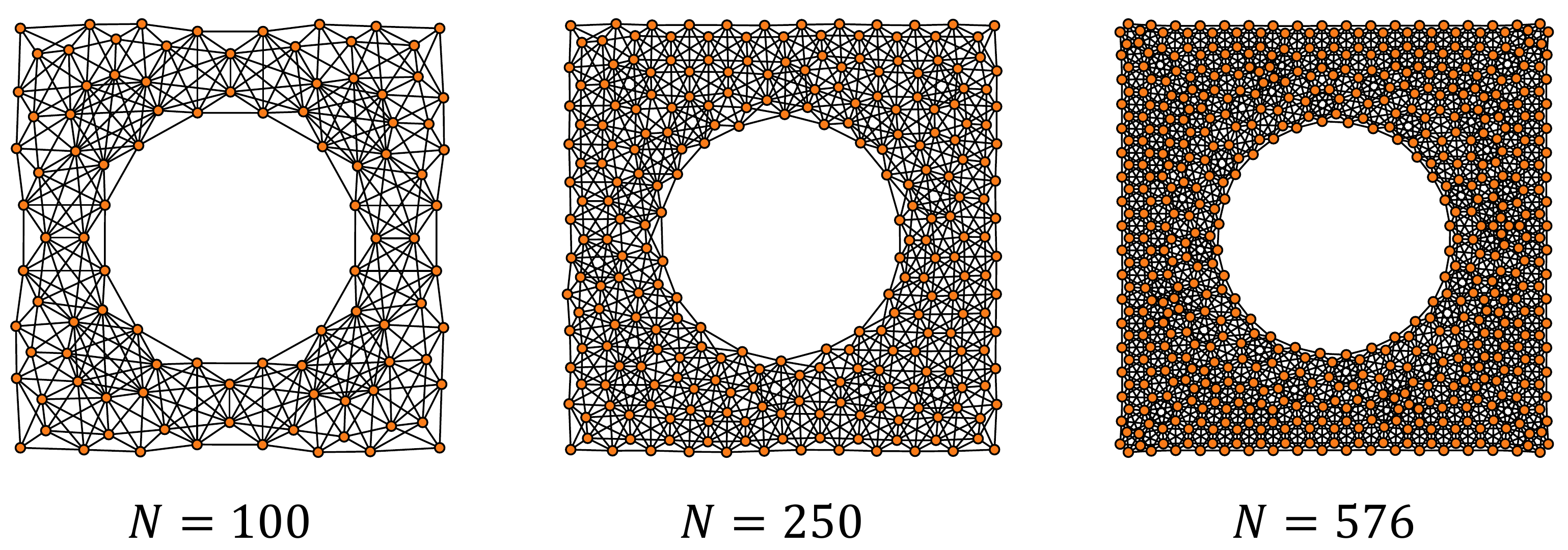}
  \label{fig:test1}
\end{minipage}%
\begin{minipage}{.5\textwidth}
  \centering
  \hspace*{1cm}  
  \vspace*{0.5cm}  
  \includegraphics[width=.7\linewidth]{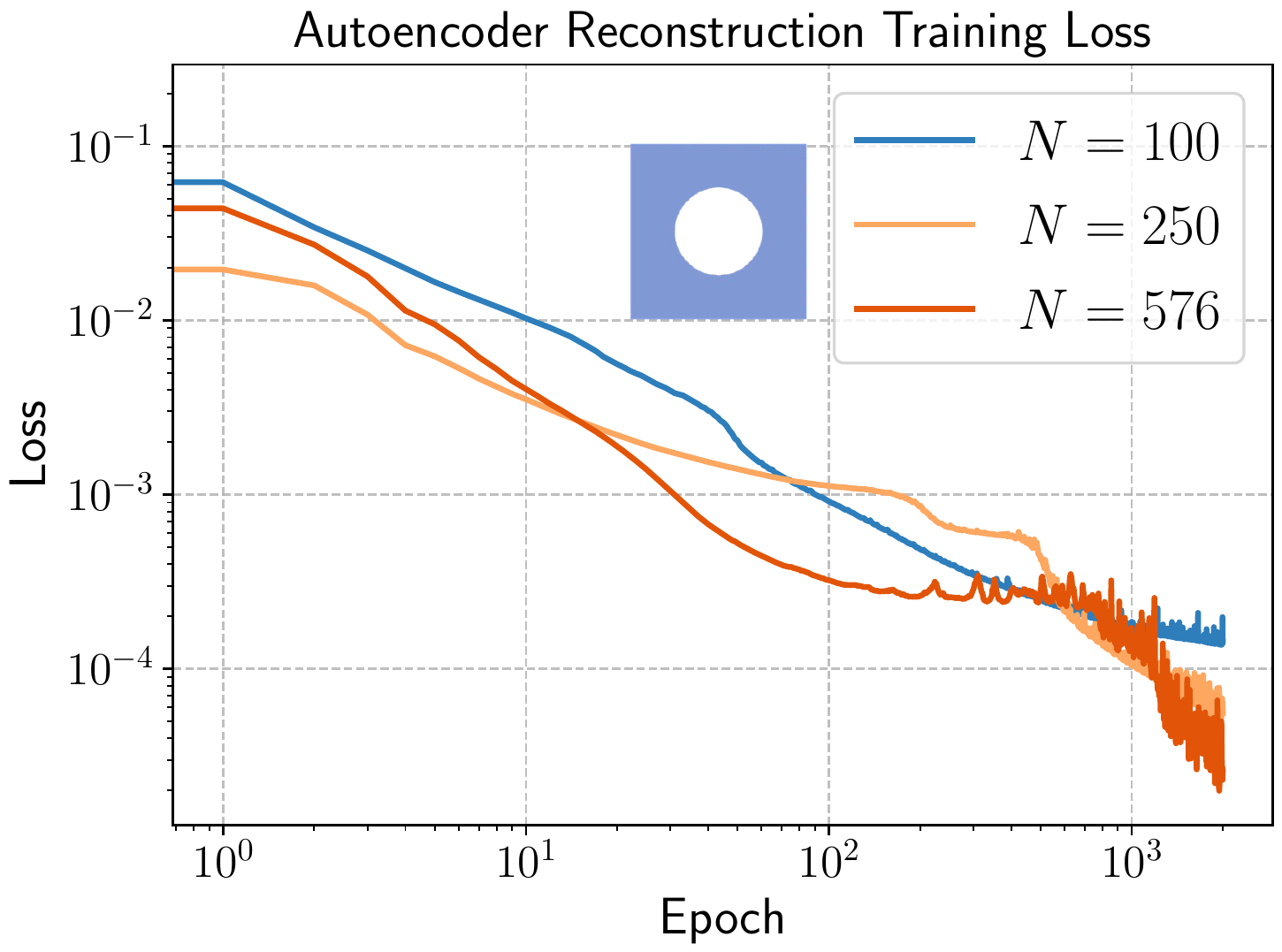}
  \label{fig:test2}
\end{minipage}
\caption{Autoencoder reconstruction training loss for microstructure A  as defined in Eq.~\ref{eq:autoencoder_loss} with the size of the encoded feature vector $D = 16$ and graph sizes of $N = 100$, $N=250$, and $N = 576$ nodes.}
\label{fig:mesh_sensitivity_autoencoder_loss}
\end{figure}

In this section, we investigate the behavior for different dimensionality of the graph data set and the compression of the graph information.
In the first experiment, we test the effect of the size of the input plasticity graph that will be reconstructed by the autoencoder.
We generate three data sets from finite element simulations with different mesh sizes for microstructure A.
All meshes consist of the same triangular elements described in the previous section.
The number of elements in the mesh and the corresponding nodes in the post-processed graphs are $N = 100$, $N = 250$, and $N=576$.
For the mesh generation, we start with the $N= 100$ mesh and refine once to obtain the $N=250$ mesh and twice to obtain the $N=576$ mesh.
The refinement was performed automatically using the meshing software library Cubit \citep{blacker1994cubit}.
The data sets for the meshes of $N= 100$, $N=250$, and $N= 576$ nodes mesh were generated through the same FEM simulation setup and a subset of the combinations of uniaxial and shear loading paths described in Section~\ref{sec:data_generation} gathering 2500 training samples of graphs.

The results of the training experiment on the mesh sensitivity are demonstrated in Fig.~\ref{fig:mesh_sensitivity_autoencoder_loss}.
The figure shows the reconstruction loss Eq.~\eqref{eq:autoencoder_loss} for the three mesh sizes.
The autoencoder architecture was identical to the one described in Section~\ref{sec:autoencoder_architecture} with the only change being the number of input nodes.
The reconstruction loss exhibits a minor improvement as the number of nodes in the graph increase.
This is attributed to the density of the information available for the autoencoder to learn the patterns from -- there is a higher resolution of adjacent nodes' features.
However, an increase in graph size may increase the duration of the training procedure. 
Since the benefit of increasing the mesh size is not significant in this set of numerical experiments, 
we opt for the $N=250$ node mesh to use for the rest of this work.

\begin{figure}[h!]
\centering
\includegraphics[width=.35\textwidth ,angle=0]{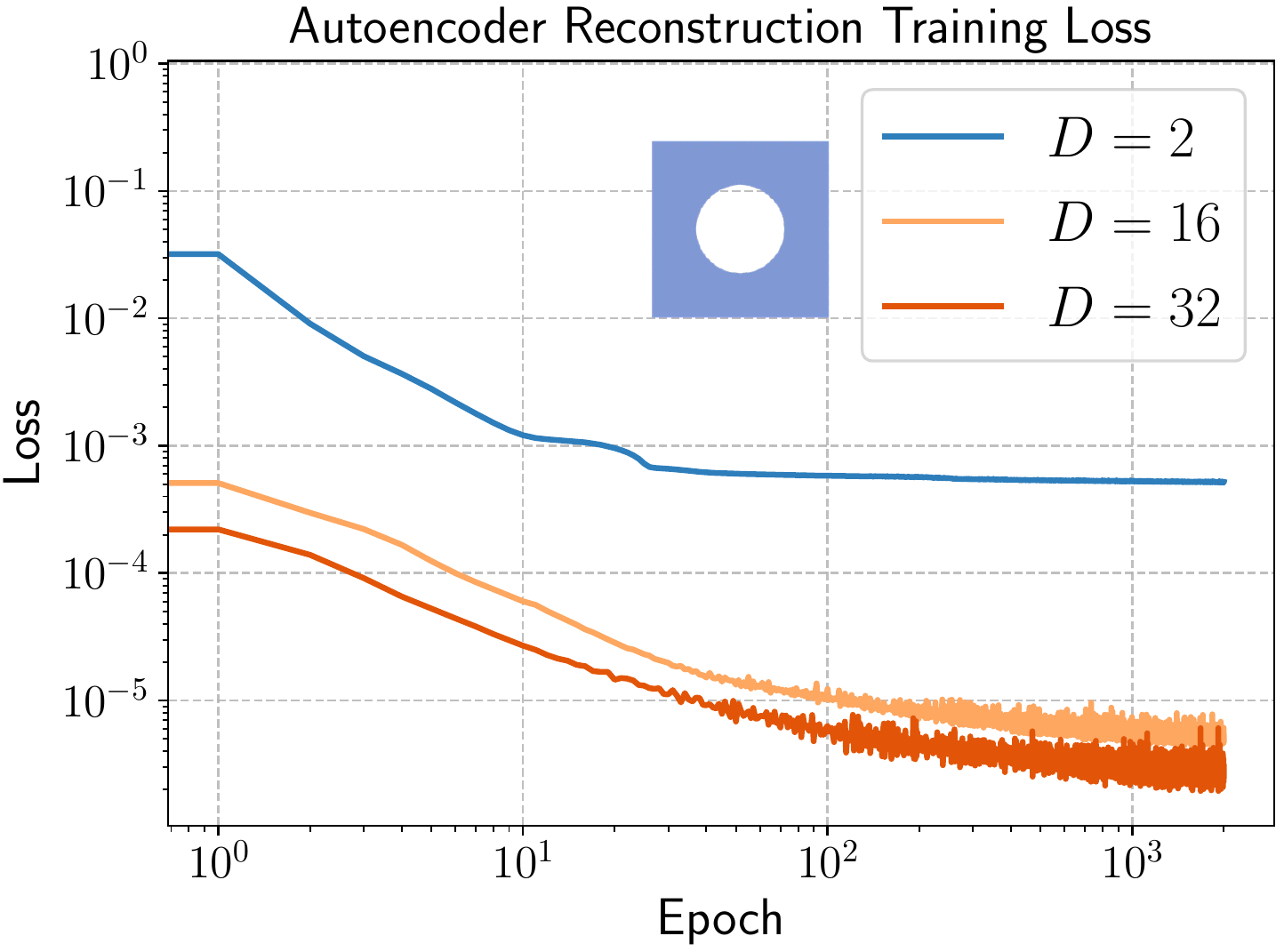} 
\caption{Autoencoder reconstruction training loss for microstructure A  as defined in Eq.~\ref{eq:autoencoder_loss} with the size of the encoded feature vector $D = 2$, $D = 16$, and $D=32$.}
\label{fig:feature_vector_size_autoencoder_loss}
\end{figure}

In a second numerical experiment, we examine the effect of the size of the encoded feature vector on the capacity of the autoencoder 
to learn and reconstruct the plasticity distribution patterns.
We re-train the autoencoder using the data set of 10000 graphs generated on the $N=250$ node mesh for microstructure A as described in Section~\ref{sec:data_generation}.
We perform three training experiments selecting different sizes $D$ for the encoded feature vector -- $D=2$, $D=16$, and $D=32$.
The autoencoder architectures are identical to the one described in Section~\ref{sec:autoencoder_architecture}. 
All the convolutional filters and the Dense layers have the same size. 
The only Dense layers that are affected are those around the encoded feature vector whose input and output sizes are modified to accommodate the different sizes of encoded feature vector.

\begin{figure}[h!]
\centering
\includegraphics[width=.90\textwidth ,angle=0]{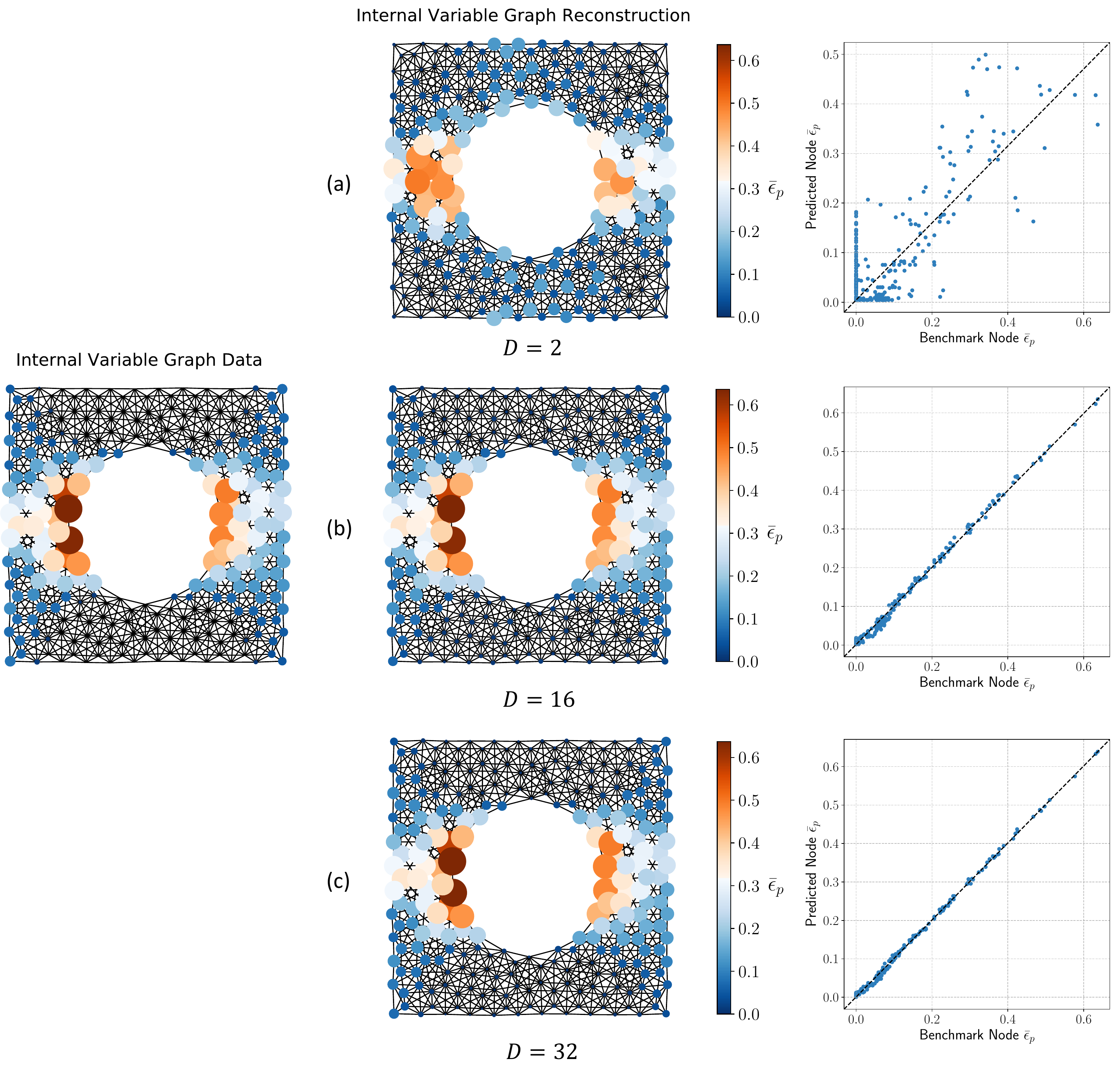} 
\caption{Comparison of the autoencoder architecture's capacity to reconstruct the accumulated plastic strain pattern of the microstructure A with the size of the encoded feature vector (a) $D = 2$, (b) $D = 16$, and (c) $D=32$. The graph node size and color represent the magnitude of the accumulated plastic strain $\overline{\epsilon}_p$. The node-wise predictions for $\overline{\epsilon}_p$ are also demonstrated. }
\label{fig:feature_vector_size_reconstruction}
\end{figure}

The training performance for these three training experiments is demonstrated in Fig.~\ref{fig:feature_vector_size_autoencoder_loss}.
Compared to encoded feature vector sizes $D=16$ and $D=32$, the $D=2$ autoencoder architecture seems to fail to compress the information as well
with a loss performance difference of about two orders of magnitude. 
The maximum compression achieved for this autoencoder architecture setup appears to be two features. This dimensionality appears to be the smallest feasible encoding limit for this particular data set. It is also possible that more sampling from different loading paths may also increase this minimal dimensionality. 
Jumping from $D=16$ to $D=32$ encoded feature vector components, only a small improvement in the reconstruction capacity is observed.
The reconstruction capacity is also illustrated in Fig.~\ref{fig:feature_vector_size_reconstruction}. 
The decoder fails to accurately reconstruct the plasticity graph from the $D=2$ encoded feature vector (Fig.~\ref{fig:feature_vector_size_reconstruction} a).
However, for $D=16$ and $D=32$ the decoder accurately reproduces the plasticity patterns (Fig.~\ref{fig:feature_vector_size_reconstruction} b and c).
It is expected for dimensions larger than $D=32$ the benefit in reconstruction capacity will me minimal.
Thus, the encoded feature vector dimension selected for the rest of this work is $D=16$ for which the dimension reduction capacity is considered adequate and computationally efficient.

\section{Numerical Example: Multiscale plasticity with graph internal variables} 
\label{sec:constitutive_predictions}
In this section, we demonstrate how the graph internal variables generated by the graph autoencoder are incorporated into the predictions at the macroscale constitutive law as well as how they can be decoded and be used to estimate and interpret the evolution of microstructures within an RVE. 
In Section~\ref{sec:constitutive_model_training}, we demonstrate the training results for the neural network constitutive models described in Section~\ref{sec:constitutive_model}.
In Section~\ref{sec:recurrent_architectures}, the predictions of these neural network constitutive models are integrated with a return mapping algorithm to make forward elastoplastic predictions and how they compare with recurrent neural network architectures from the literature.
In Section~\ref{sec:graph_interpretation}, we show the forward prediction capacity of the neural network models in unknown loading paths and demonstrate the behavior of the encoded feature vector prediction variables and how they can be translated back to the original graph space with the help of the graph decoder.

\subsection{Training of constitutive models}
\label{sec:constitutive_model_training}

\begin{figure}[h!]
\centering
\includegraphics[width=.65\textwidth ,angle=0]{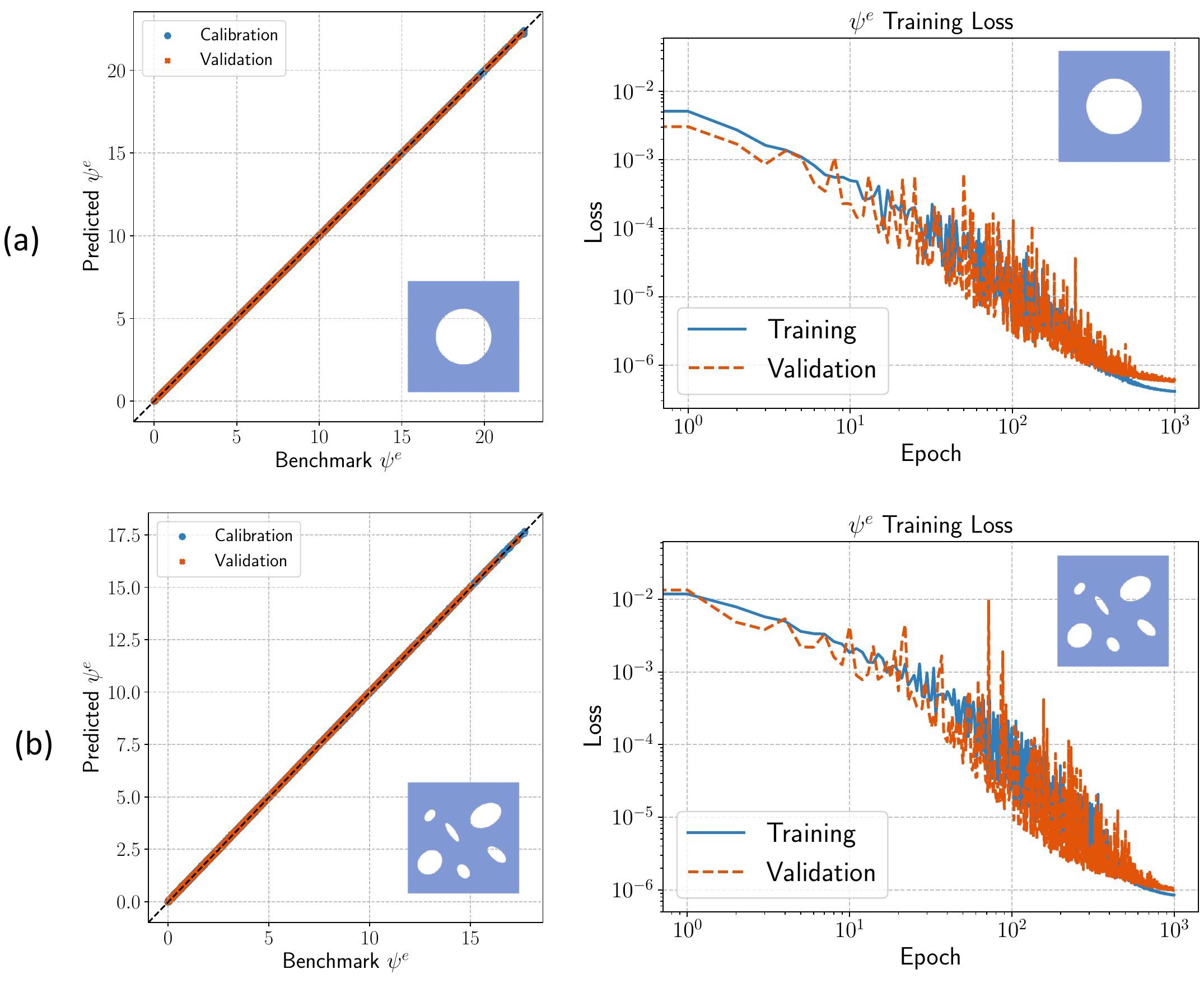} 
\caption{Prediction and training loss of the hyperelastic energy functional $\psi^{e}$ constitutive models for microstructures (a) A and (b) B. }
\label{fig:elasticity_training}
\end{figure}
In this section, we demonstrate all the training experiments for the neural network components that will be used in the return mapping algorithm to make forward elastoplastic predictions as described in Section~\ref{sec:constitutive_model}. The training experiments for the hyperalstic energy functional, the yield function, and the kinetic law are performed for both microstructures A and B. 

In a first training experiment, we first validate the capacity of capturing the hyperelastic behavior of the material in different loading directions.
Neural network hyperelastic laws has been previously trained to be compatible with mechanics knowledge (e.g. ployconvexity) and thermodynamic constraints
\citep{le2015computational,klein2022polyconvex, vlassis2022molecular}.
The feed-forward neural network inputs the elastic strain tensor in Voigt notation in two dimensions ($\epsilon_{11},\epsilon_{22},\epsilon_{12}$) and outputs the hyperelastic energy function $\widehat{\psi}^e$.
Through differentiation of the network's prediction, the stress and stiffness are also output and are constrained with a higher-order Sobolev norm similarly formulated as in \cite{vlassis2021sobolev} that is omitted for brevity.
The hyperelastic network was trained and validated for each microstructure on a total of 10000 sample points (8000 used for training and 2000 for validation) that were gathered from the FEM simulations described in the previous section.
The hyperelastic constitutive response was recorded in both the elastic and plastic regions. During the plastic response, the elastic increment that corresponds to the true plastic increment is recorded.

\begin{figure}[h!]
\centering
\includegraphics[width=0.99\textwidth ,angle=0]{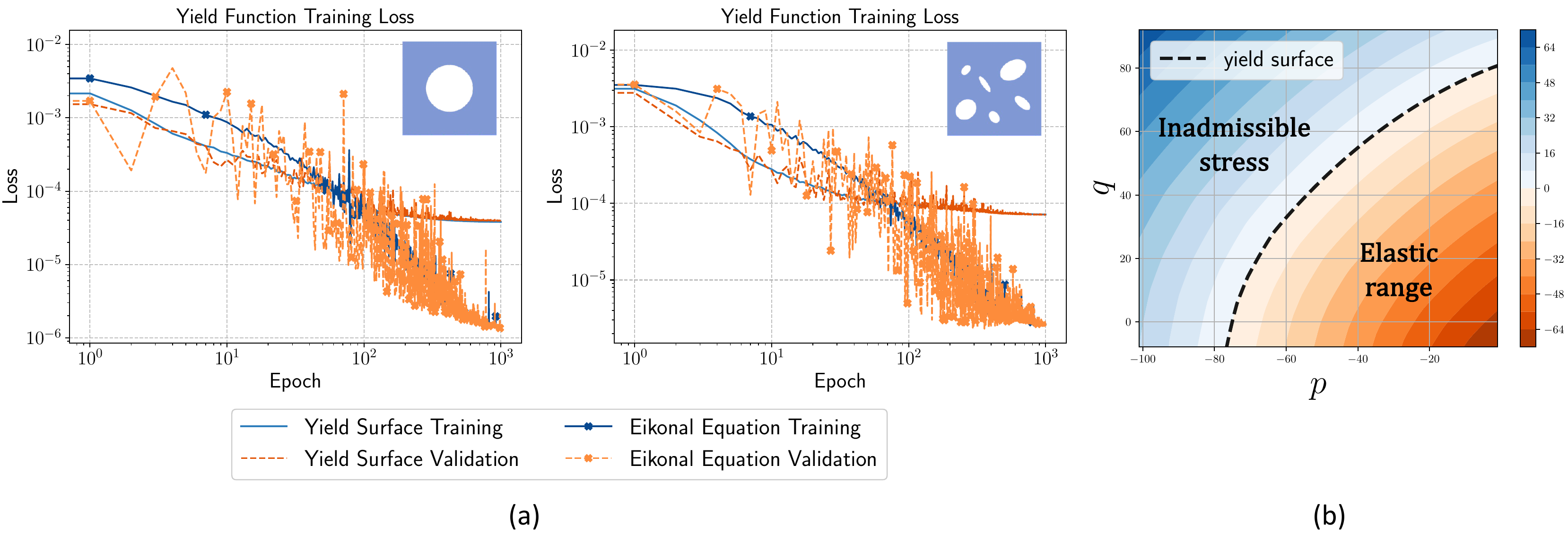} 
\caption{(a) Training loss curves for the yield surface and Eikonal equation loss terms (Eq.~\eqref{eq:yield_function_loss}) for microstructures A and B. 
(b) An instance of the yield function level set prediction for microstructure A. }
\label{fig:yield_function_training}
\end{figure}

The neural network follows a multilayer perceptron architecture.
It consists of a hidden Dense layer (100 neurons/ReLU), followed by one Multiply
layer, then another hidden Dense layer (100 neurons/ReLU), and an output Dense layer (Linear). 
The Multiply layer was first introduced in \citep{vlassis2021sobolev} to modify and increase the degree of continuity of the neural network's hidden layers.
It performs a simple elementwise multiplication of a layer's output with itself.
It was shown to increase the smoothness of the learned functions and also allow for better control and reduction of the higher-order constraints, such as a $H_2$ loss function terms.
The layers’ kernel
weight matrix was initialized with a Glorot uniform distribution and the bias vector with a zero distribution.
The model was trained for 1000 epochs with a batch size of 100 using the Nadam optimizer, set with default values.
The training curves and the predictions for all the samples in the data set are demonstrated in Fig.~\ref{fig:elasticity_training} for which equally great performance was observed for both microstructures.

\begin{figure}[h!]
\centering
\includegraphics[width=0.99\textwidth ,angle=0]{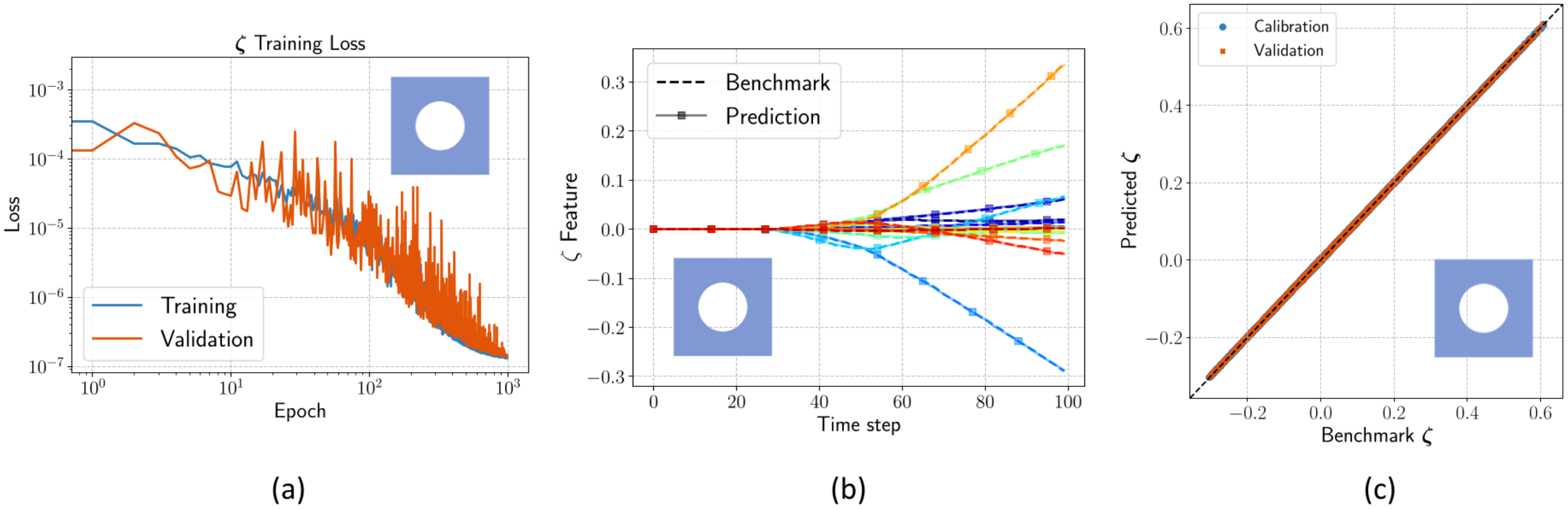} 
\includegraphics[width=0.99\textwidth ,angle=0]{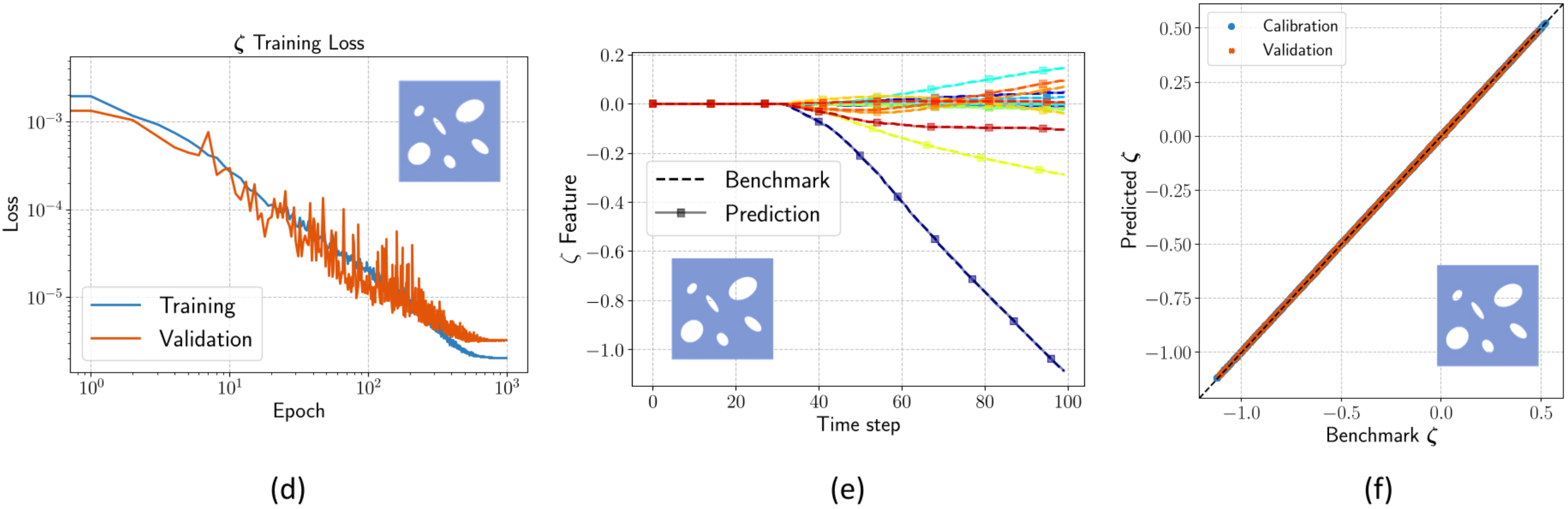} 
\caption{(a,d) Training loss curves for the encoded feature vector $\widehat{\tensor{\zeta}}$ kinetic law, (b,e) a prediction of the encoded feature vectors along a loading path, and (c,f) prediction of all the encoded feature vector components in the data set for microstructures A and B. }
\label{fig:zeta_network_training}
\end{figure}

\begin{figure}[h!]
\centering
\includegraphics[width=0.75\textwidth ,angle=0]{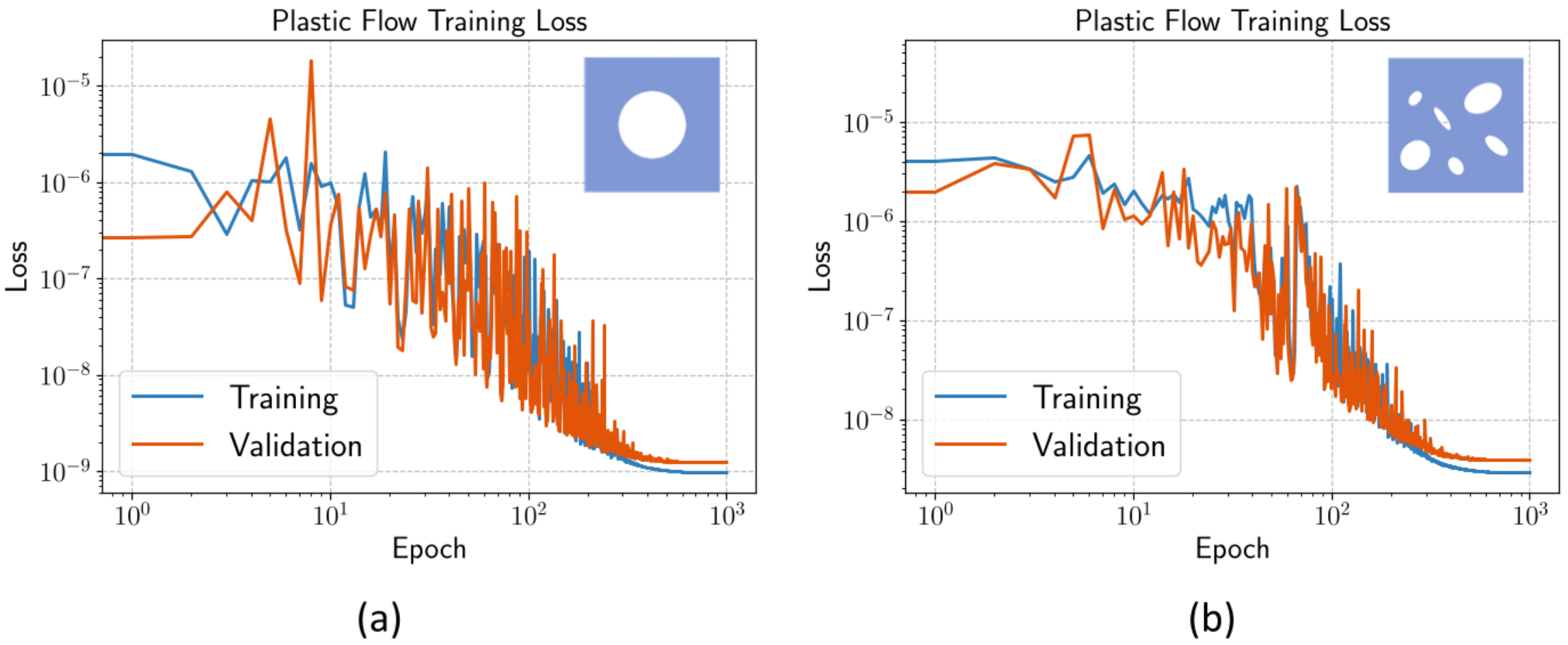} 
\caption{Training curves for the plastic flow network for microstructures A and B. }
\label{fig:flow_network_training}
\end{figure}

Similarly, the neural network yield function is trained on a total of 10000 sample points (8000 and 2000 for training and validation respectively) for each microstructure.
The formulation and training objective for the feed-forward network is as described in Section~\ref{sec:yield_function}.
It consists of a hidden Dense layer (100 neurons/ReLU), followed by a Multiply
layer, then another hidden Dense layer (100 neurons/ReLU) and another Multiply layer, and the output is fed in a Dense layer with a linear activation function. 
For all the Dense layers, the kernel
weight matrix was initialized with a Glorot uniform distribution and the bias vector with a zero distribution.
Each model was trained for 1000 epochs with a batch size of 100 with the Nadam optimizer, set with default values.
The Eikonal equation term weight in the loss function Eq.~\eqref{eq:yield_function_loss} is set to $w=1$.
The training results for the yield function neural networks are showcased in Fig.~\ref{fig:yield_function_training}.
The training loss curves for the two terms of the training loss function in Eq.~\eqref{eq:yield_function_loss} are shown in Fig.~\ref{fig:yield_function_training}(a) with similar performance for both microstructures.
The Eikonal equation loss term aims to ensure the predicted yield function is a signed distance function and a predicted instance of it is demonstrated in Fig.~\ref{fig:yield_function_training}(b).

We fit the kinetic law neural networks as described in Section~\ref{sec:kinetic_law} to predict the encoded feature vectors as a function of the plastic strain tensor time history.
The architecture was trained for each data set of 10000 samples of plastic strain and encoded feature vector pairs, split into an 8000 training and a 2000 validation set.
The plastic strain tensors were pre-processed in time history sequences of length $\ell = 4$.
The recurrent neural network architecture is based on the Gated Recurrent Unit-based (GRU) \citep{chung2014empirical}. 
The network consists of two GRU hidden layers of 32 units each with a sigmoid recurrent activation function and a tanh layer activation function).
This is followed by two Dense hidden layers (100 neurons and a ReLU activation function) and an output Dense layer (16 neurons and a Linear activation function).
The model was trained for 1000 epochs with a batch size of 128 using the Nadam optimizer, set with default values, and the training curves and encoded feature vector components prediction are showcased in Fig.~\ref{fig:zeta_network_training}.

In the last training experiment, we train the plastic flow neural networks to predict the plastic flow evolution in the elastoplasticity simulations as described in Section~\ref{sec:plastic_flow}.
The architecture was trained for each data set of 10000 samples of encoded feature vector and plastic flow component pairs, split into an 8000 training and a 2000 validation set.
The input encoded feature vector has 16 components.
The multilayer perceptron architecture consists of four hidden Dense layers of 100 neurons each and a ReLU activation function.
The output of the neural network is a Dense layer with 2 neurons and a Linear activation function.
The layers’ kernel weight matrix was initialized with a Glorot uniform distribution and the bias vector with a zero distribution.
The model was trained for 1000 epochs with a batch size of 100 using the Nadam optimizer.
The training curves for the plastic flow networks for the two microstructures are demonstrated in Fig.~\ref{fig:flow_network_training}.
As expected these neural networks are very accurate as they input the highly descriptive encoded feature vector structure that represents the entire plastic state distribution in the microscale
to predict the homogenized plastic flow behavior.

\subsection{Comparison with recurrent neural network architectures}
\label{sec:recurrent_architectures}

In this section, we test the capacity and robustness of the return mapping elastoplasticity model to make forward path-dependent predictions on unseen loading paths.
We compare this capacity with recurrent neural network models from the literature that are commonly used to predict data structures in the form of time series and often plasticity.
We design training experiments for the two recurrent architectures: a GRU architecture and a 1D convolutional architecture and test all models against unseen loading paths of increased complexity.

\begin{figure}[h!]
\centering
\includegraphics[width=0.99\textwidth ,angle=0]{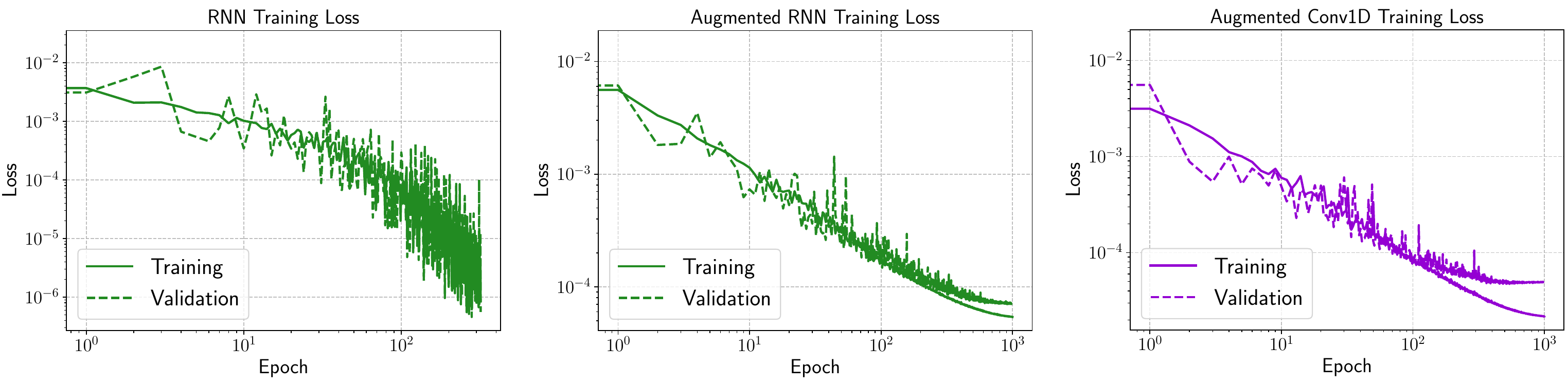} 
\caption{Training loss curves for the RNN architecture, RNN architecture with data augmentation, and the Conv1D architecture with data augmentation. }
\label{fig:rnn_loss}
\end{figure}

The first recurrent network used is based on the GRU layer architecture that is trained on a pair of total strain and total stress time histories.
For this 2D data set, the strain state is represented by the strain tensor $\tensor{\epsilon}$ in Voigt notation $(\epsilon_{11},\epsilon_{22},\epsilon_{12})$, and the stress state is represented by the two stress invariants $(p,q)$.
Specifically, the network inputs a time history of input strains of $\ell$ previous time steps to predict the current stress state. 
For the $n$-th time step, the network inputs the pre-processed time history of strain tensors as $\left[\tensor{\epsilon}_{n-\ell}, \ldots, \tensor{\epsilon}_{n-1}, \tensor{\epsilon}_{n}\right]$ and outputs the current stress $\left[ p_n,q_n \right]$.
The time history length was chosen to be $\ell = 30$.
The recurrent neural network architecture (RNN) consists of a series of two GRU hidden layers (32 units each, a sigmoid recurrent activation function, and a tanh activation function) and a series of two Dense hidden layers (100 neurons and a ReLU activation function) with an output Dense layer (2 neurons and a Linear activation function).
The model was trained for 1000 epochs with a batch size of 128 using the Nadam optimizer, set with default values, and a mean squared error loss function.

\begin{figure}
\centering
\hspace*{-0.5cm}  
\begin{minipage}{.53\textwidth}
  \centering
  \includegraphics[width=1\linewidth]{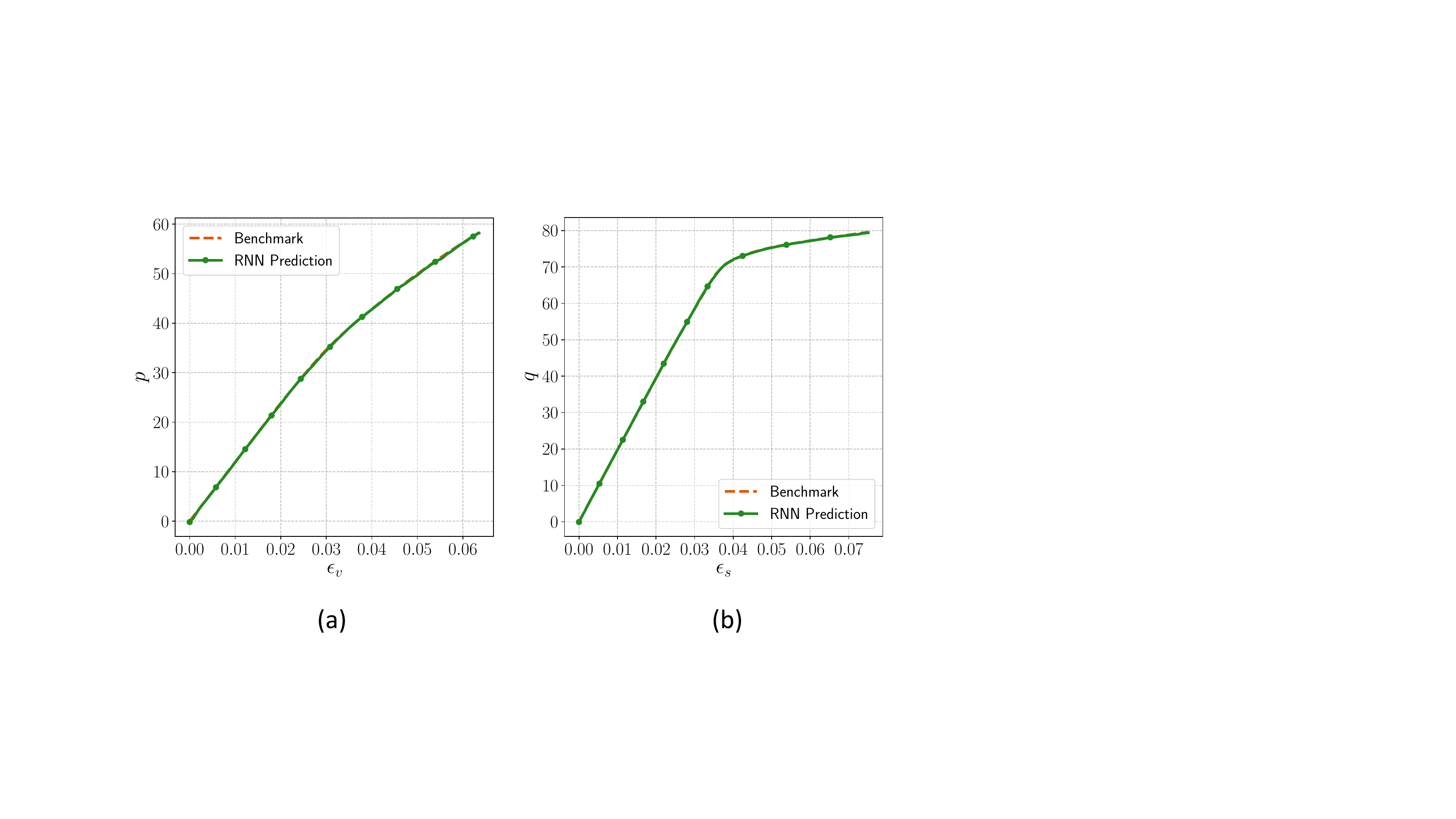}
  \label{fig:test1}
\end{minipage}%
\begin{minipage}{.53\textwidth}
  \centering
  \includegraphics[width=1\linewidth]{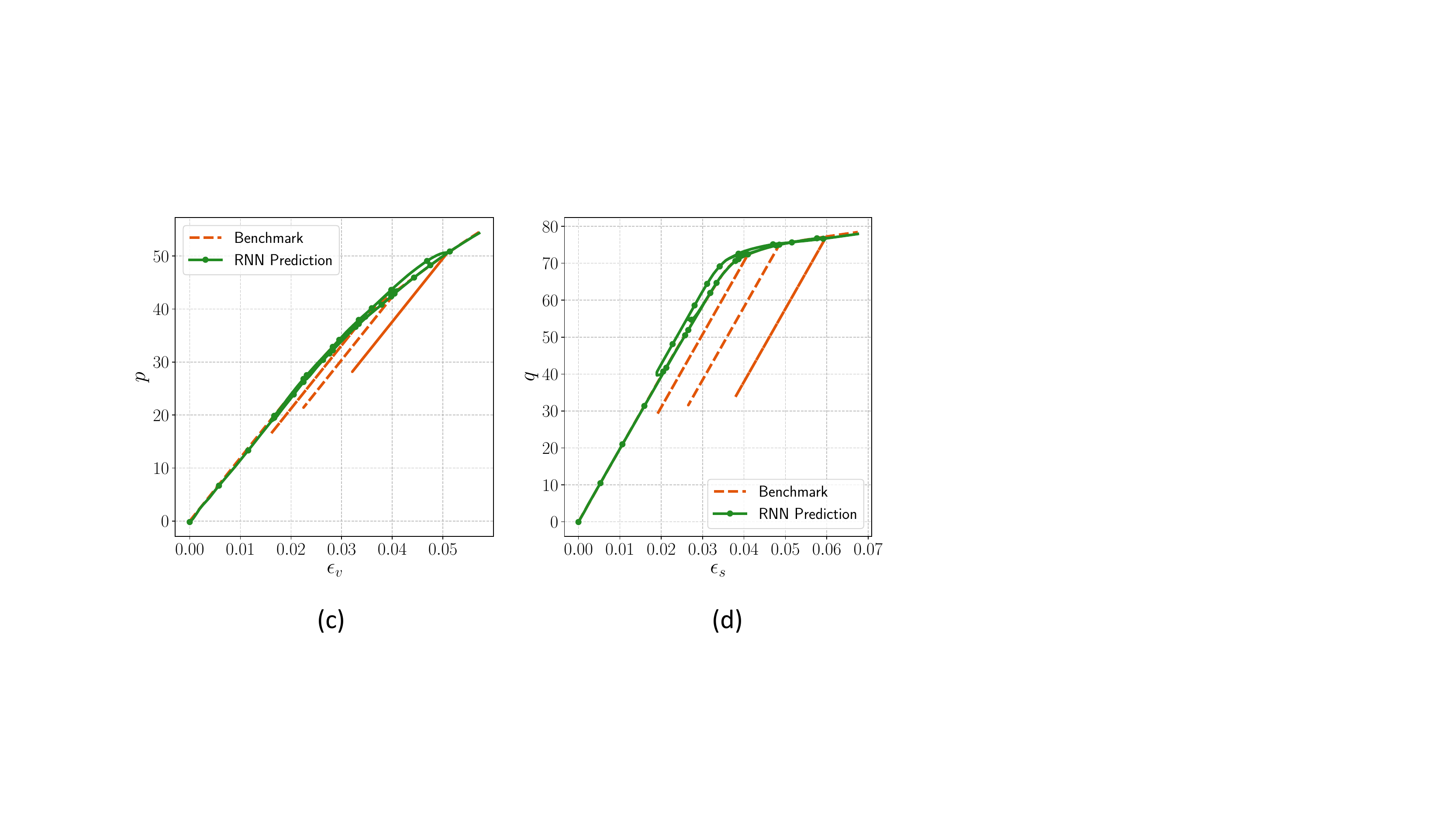}
  \label{fig:test2}
\end{minipage}
\caption{(a,b) Prediction of the RNN architecture on a monotonic loading curve. (c,d) Prediction of the RNN architecture for a loading curve with unseen random unloading and reloading paths. }
\label{fig:rnn_mono}
\end{figure}

The second recurrent architecture used is based on the 1D convolutional layer architecture that uses a one-dimensional variation of the convolutional filter \citep{lecun1995convolutional,oord2016wavenet} to extract information from time series of fixed length. The plasticity data time series is similarly pre-processed and input as the previously described RNN architecture with a time history length $\ell = 30$.
The 1D convolutional neural network architecture (Conv1D) consists of a series of three 1D convolutional layers with 32, 64, and 128 filters respectively, with a kernel size of 4 with ReLU activation functions.
The output of the convolutional filter is flattened and fed into two Dense layers (100 neurons and ReLU activation functions) and an output Dense layer (2 neurons and a Linear activation).
The model was trained for 100 epochs with a batch size of 128, the Nadam optimizer, and a mean squared error loss function.

In a first training experiment, before we compare with the neural network return mapping model, we train the RNN architecture with the same data set for microstructure A that the return mapping constitutive model was trained on as described in Section~\ref{sec:data_generation}.
It is noted that all the loading paths in that data set were monotonic. The 10000 samples of strain-stress constitutive responses were pre-processed in 10000 strain time history samples and their corresponding stress response -- 8000 were used for training and 2000 were used for calibration. The training loss curve for the RNN architecture is shown in Fig.~\ref{fig:rnn_loss}.
To test the capacity of the architecture to make forward predictions, we first test the RNN in unseen monotonic loading paths.
The results can be seen in Fig.~\ref{fig:rnn_mono} (a \& b) where the RNN architecture can robustly predict monotonic loading patterns that resemble the ones used for training.
However, when we introduce several random unloading and reloading paths Fig.~\ref{fig:rnn_mono} (c \& d), the RNN architecture cannot make predictions which is expected as it has not seen these types of patterns before in the training data set.

\begin{figure}
\centering
\hspace*{-0.5cm}  
\begin{minipage}{.53\textwidth}
  \centering
\includegraphics[width=1\textwidth ,angle=0]{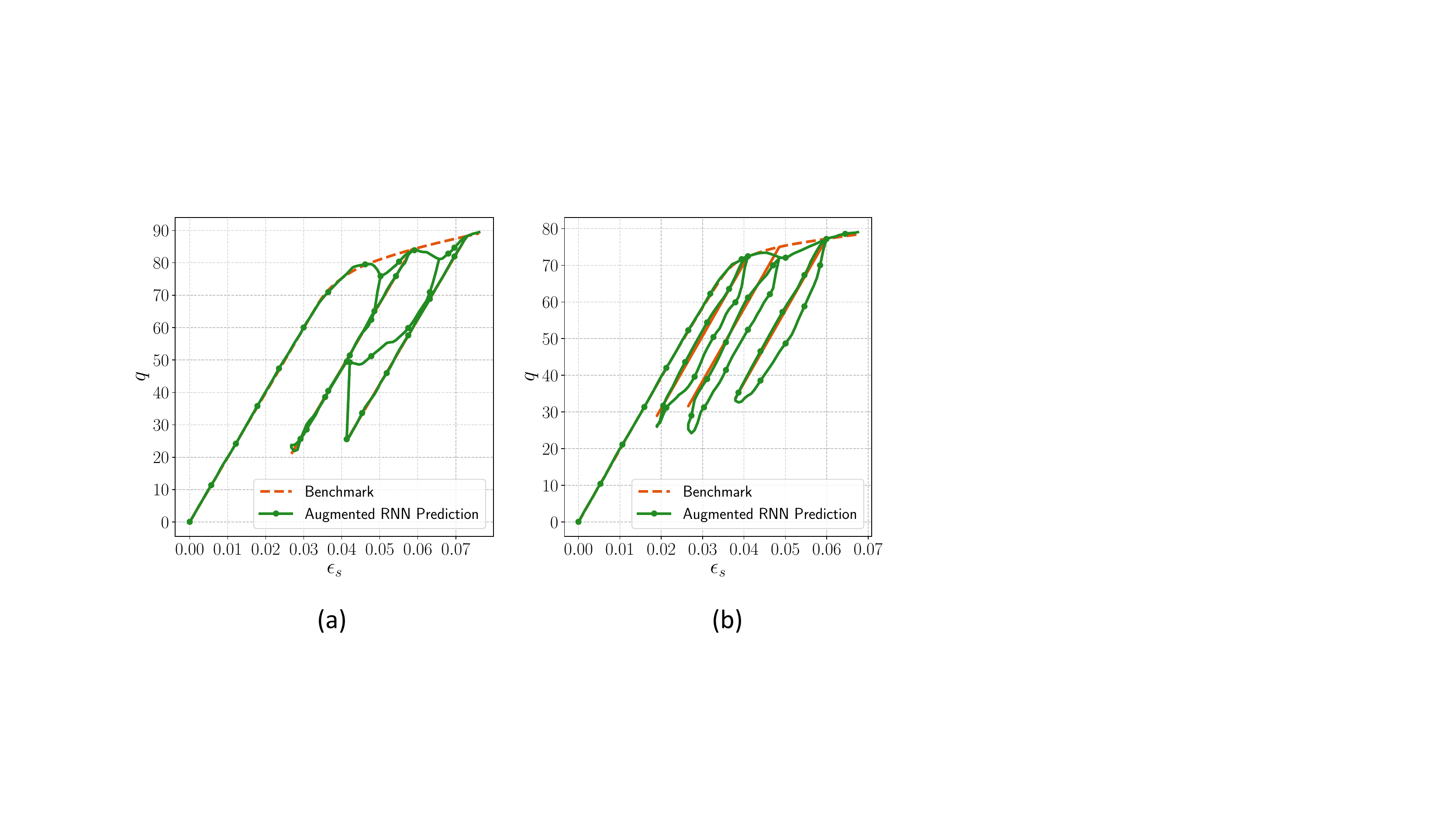} 
  \label{fig:test1}
\end{minipage}%
\begin{minipage}{.53\textwidth}
  \centering
\includegraphics[width=1\textwidth ,angle=0]{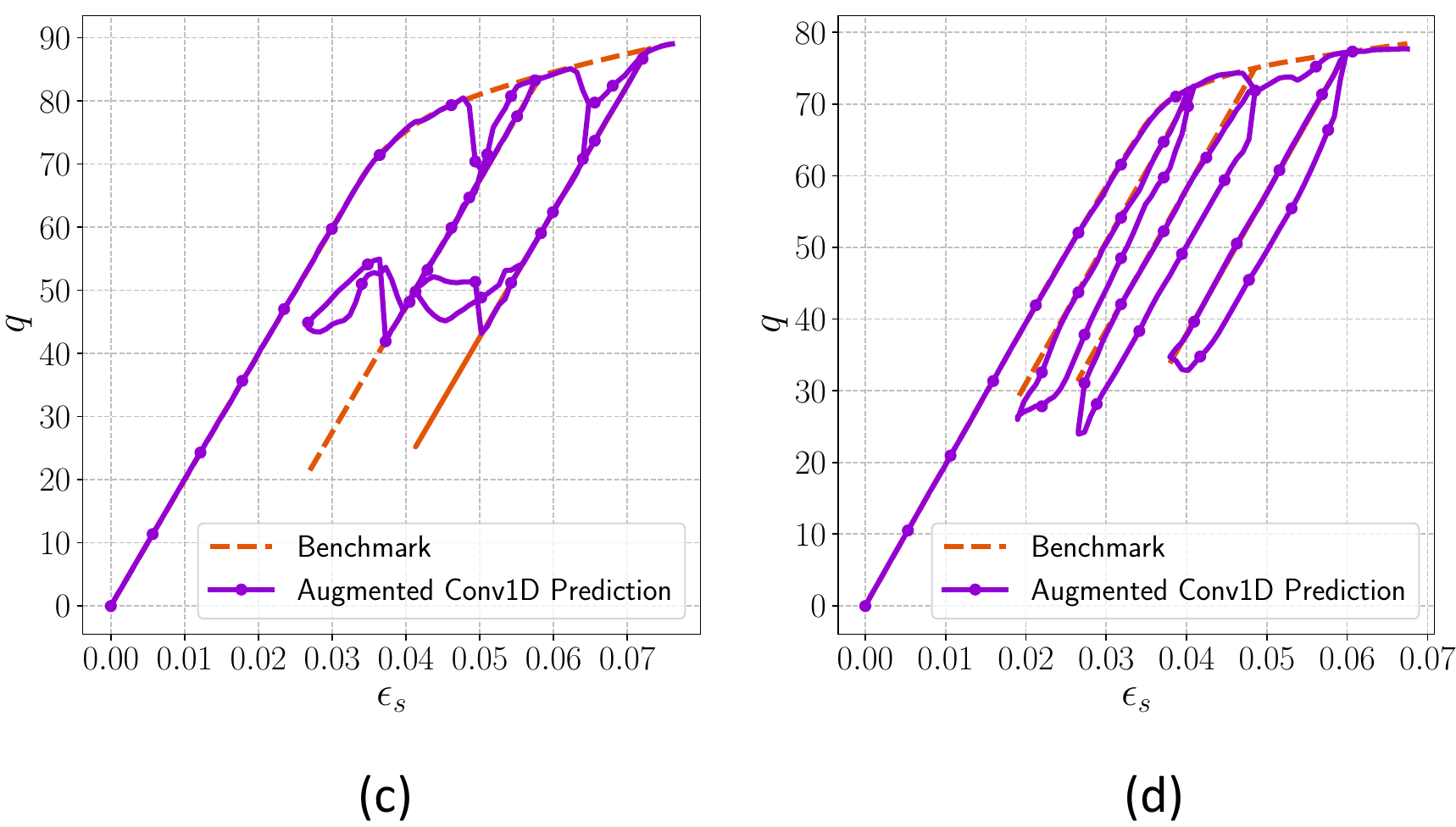} 
  \label{fig:test2}
\end{minipage}
\includegraphics[width=0.53\textwidth ,angle=0]{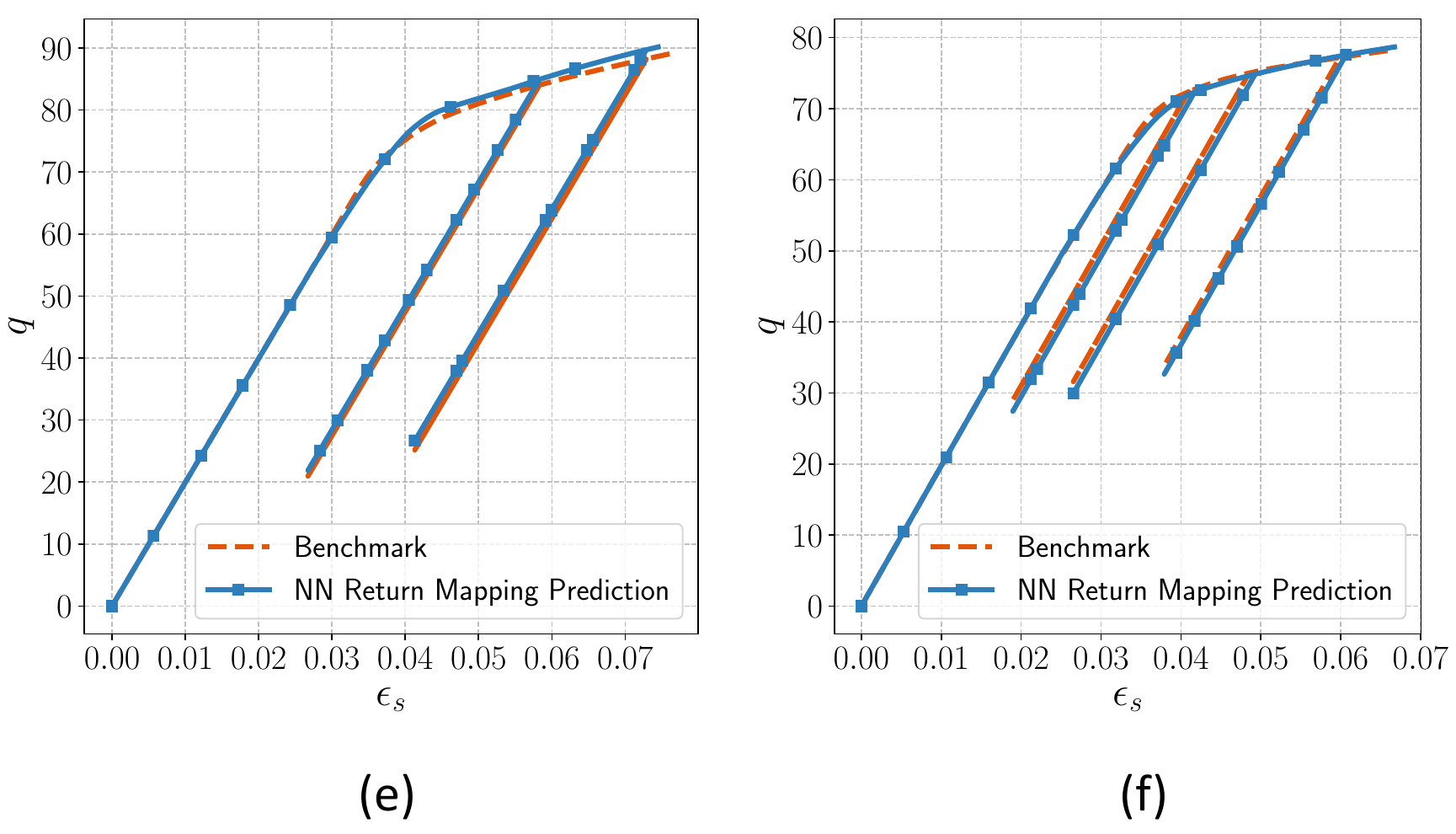} 

\caption{Prediction of the data augmented (a,b) RNN architecture, (c,d) Conv1D architecture, and (e,f) the neural network-based return mapping algorithm for loading curves with unseen random unloading and reloading paths.}
\label{fig:rnn_multi}
\end{figure}

Thus, we design another training experiment for the recurrent architectures to compare more fairly with the return mapping algorithm.
We augment the training data set for the RNN and Conv1D architectures by introducing random elastic loading and unloading paths of random lengths to the previously monotonic data set.
The data set now includes 32400 samples of strain and stress responses -- the previous 10000 samples and 22400 samples generated through the augmentation procedure.
25920 of these samples will be used for training and 6480 for validation. The training curves for the Augmented RNN and Conv1D architectures are shown in Fig.~\ref{fig:rnn_loss}.

We can now compare these augmented architectures with the neural network return mapping algorithm.
It is noted that all the return mapping algorithm models were still trained on the 10000 sample data set as described in Sec.~\ref{sec:constitutive_model_training}.
The results of the comparison experiment for unseen loading paths are demonstrated in Fig.~\ref{fig:rnn_multi}.
In Fig.~\ref{fig:rnn_multi} (a, b, c, d), the augmented RNN and Conv1D architectures can now recognize loading and reloading behaviors qualitatively better.
While several unseen unloading-reloading paths are more accurate than others the recurrent architectures are not robust in distinguishing the elastic from the plastic region.
In some elastic paths, we observe path-dependent phenomena that are potentially attributed to the network not seeing an elastic region at the specific strain states.
The return mapping algorithm model is observed to be both accurate and robust in these predictions (Fig.~\ref{fig:rnn_multi} (e \& f)), even when trained to about a third of the data compared to recurrent architectures.
This is achieved by decoupling the elastic and plastic behavior with the help of the yield function neural network -- allowing the model to recognize when the elastic and elastoplastic behaviors should be used for the predictions.

Another important point is that the black-box recurrent neural networks do not allow access to other plasticity metrics during the forward prediction simulations.
Unless there is another network training to predict metrics such as the accumulated plastic strain or other descriptors that evolve in the plastic region, we neither monitor nor can we estimate how the microstructure evolves upon yielding. 
In the following section, we will demonstrate how these feature vectors can be interpreted with the help of the graph decoder to have a complete understanding of the predicted plastic state distribution within the RVE.

\subsection{Interpretable multiscale plasticity of complex microstructures}
\label{sec:graph_interpretation}

\begin{figure}[h!]
\centering
\includegraphics[width=0.75\textwidth ,angle=0]{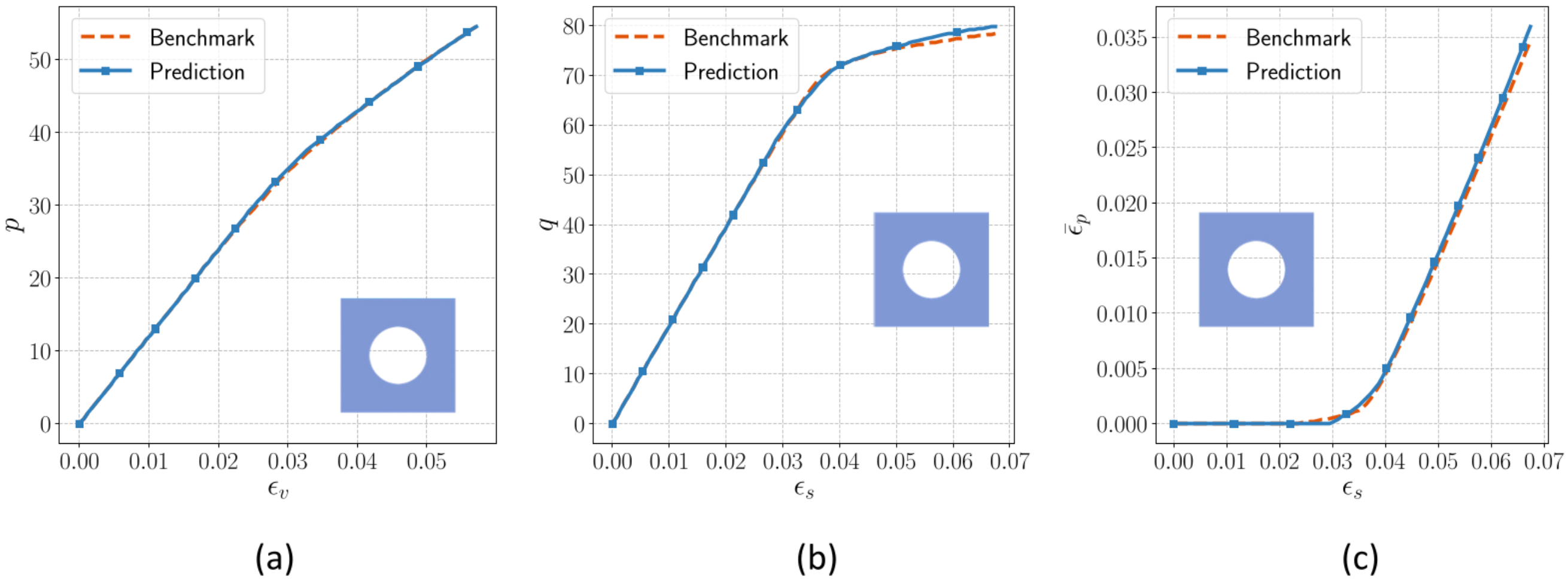} 
\includegraphics[width=0.85\textwidth ,angle=0]{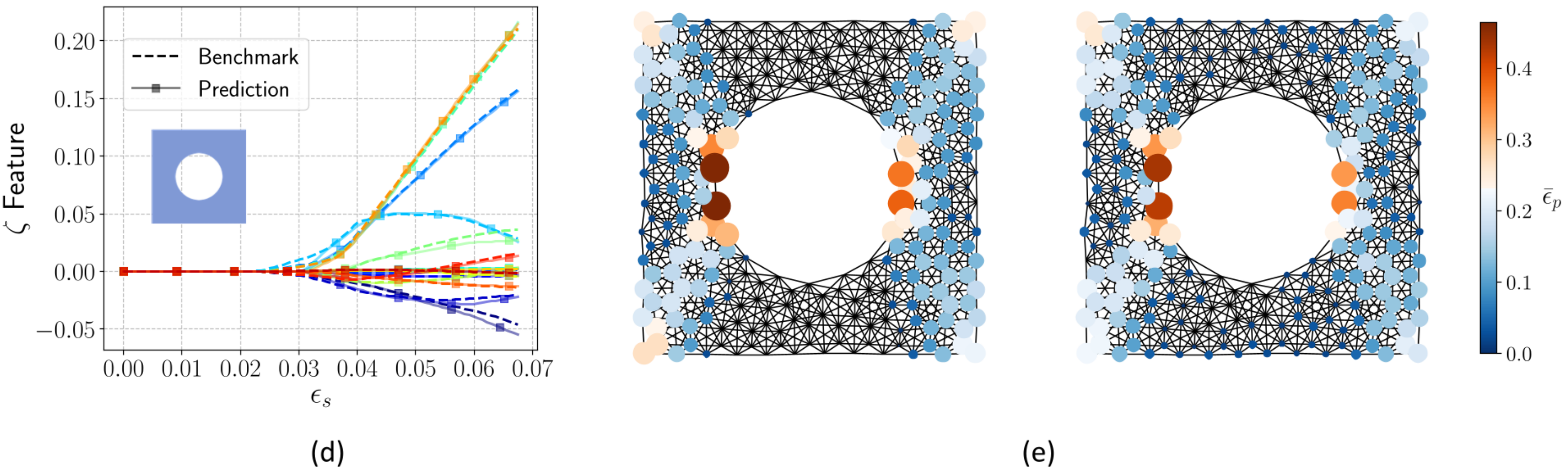} 

\caption{(a,b,c) Prediction of stress invariants $p$, $q$, and accumulated plastic strain $\overline{\epsilon}_p$ using the return mapping algorithm for a monotonic loading of microstructure A.
(d,e) Prediction of all the encoded feature vector $\tensor{\zeta}$ components and the corresponding decoded internal variable graph for a monotonic loading of microstructure A.  }
\label{fig:monotonic_loading_A}
\end{figure}

\begin{figure}[h!]
\centering
\includegraphics[width=0.75\textwidth ,angle=0]{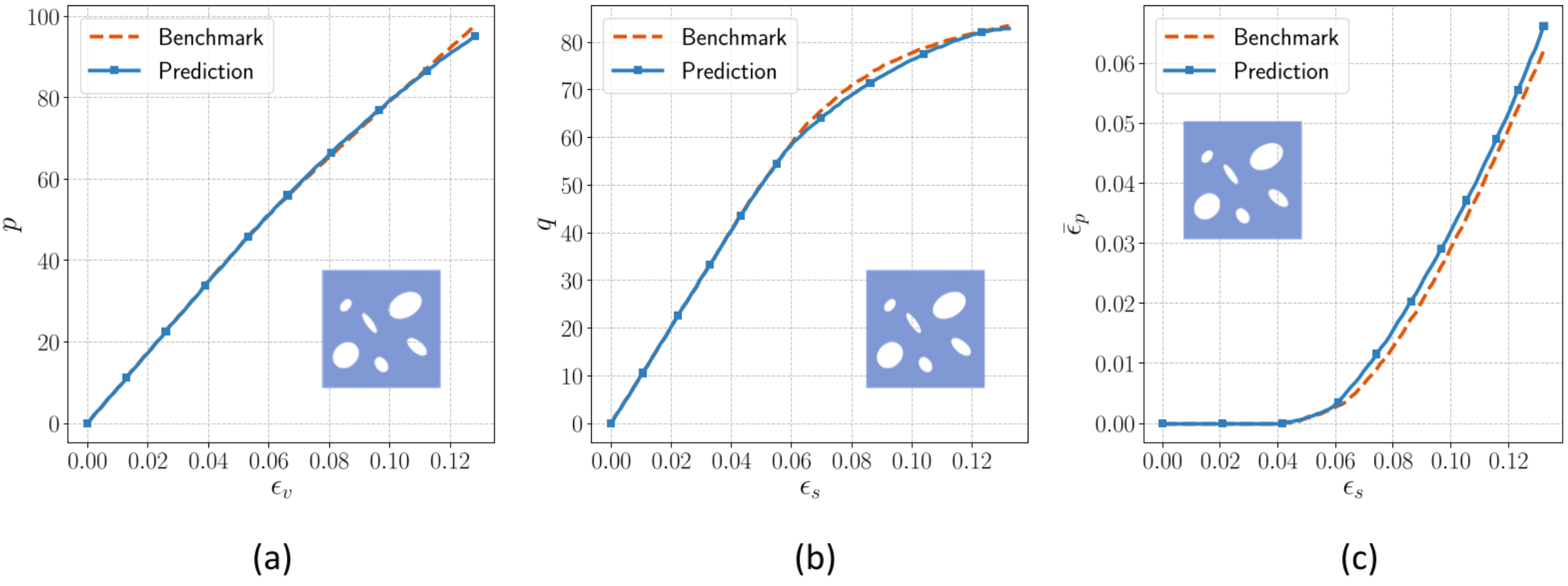} 
\includegraphics[width=0.85\textwidth ,angle=0]{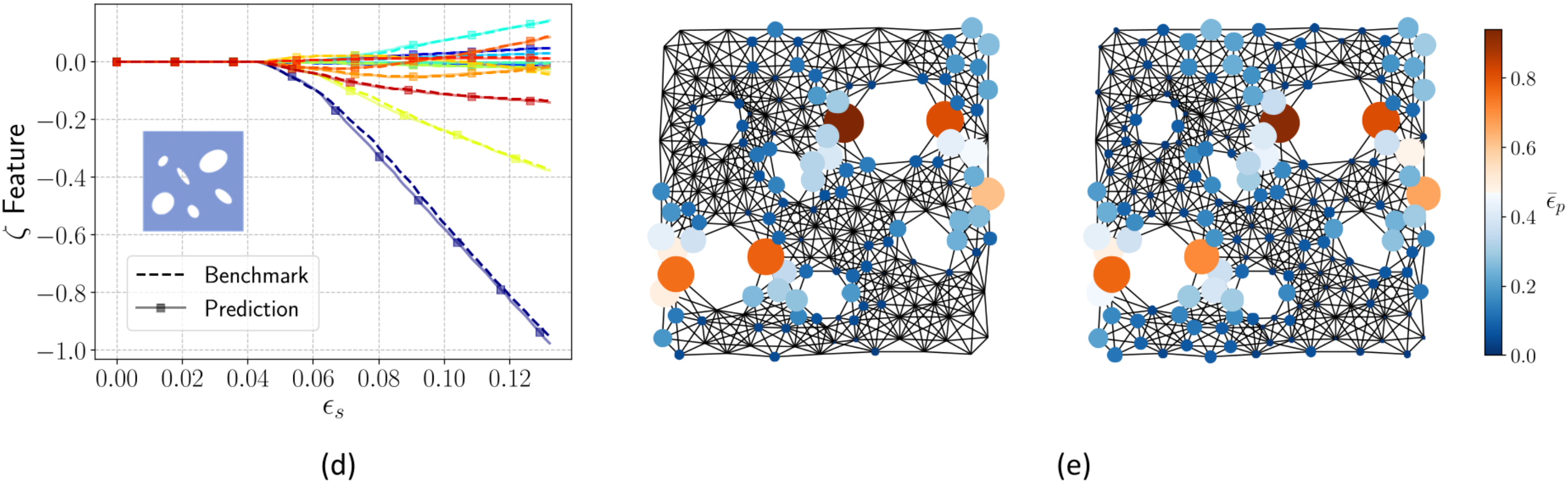} 

\caption{(a,b,c) Prediction of stress invariants $p$, $q$, and accumulated plastic strain $\overline{\epsilon}_p$ using the return mapping algorithm for a monotonic loading of microstructure B.
(d,e) Prediction of all the encoded feature vector  $\tensor{\zeta}$ components and the corresponding decoded internal variable graph for a monotonic loading of microstructure B.  }
\label{fig:monotonic_loading_B}
\end{figure}

\begin{figure}[h!]
\centering
\includegraphics[width=0.99\textwidth ,angle=0]{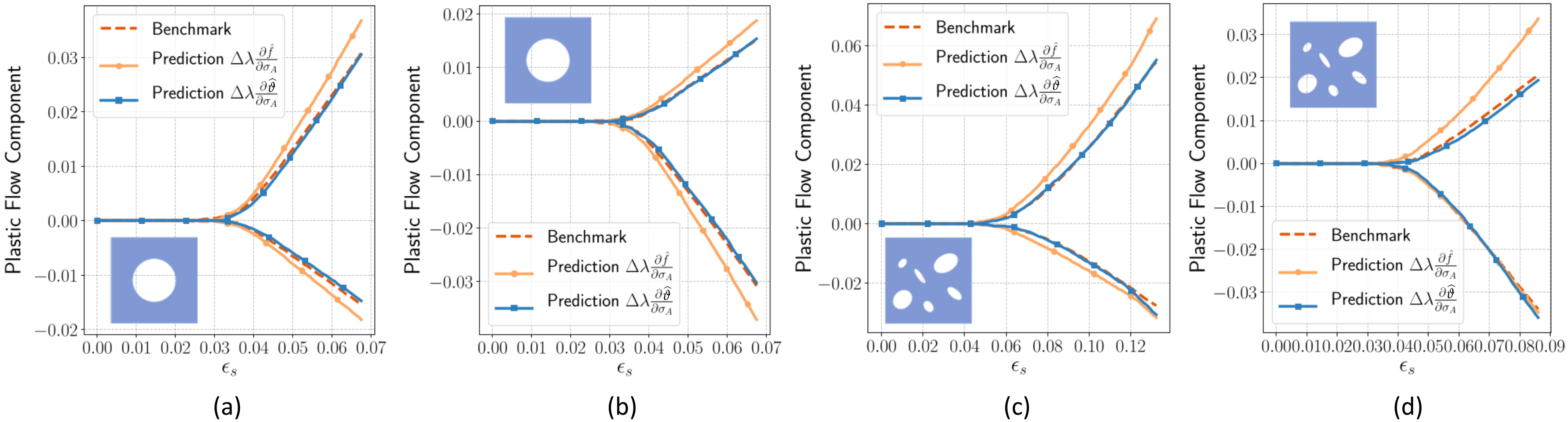} 
\caption{Prediction of the plastic flow components for two loading cases for microstructures A and B (a \& b respectively).
Both the predicted via  yield function stress gradient $\frac{\partial\widehat{f}}{\partial\sigma_A}$ and the non-associative plastic flow $\frac{\partial\widehat{\vartheta}}{\partial\sigma_A}$ predictions are shown. }
\label{fig:flow_prediction}
\end{figure}

In this last section, we demonstrate the capacity of the models to make forward predictions for unseen loading paths and interpret them in the microstructure.
The return mapping algorithm does not only predict the strain-stress response of the material but also the plastic strain response and the encoded feature vector variables.
These can then be decoded by the graph decoder $\mathcal{L}_{\text{dec}}$ to interpret the microstructures' elastoplastic behavior.
We provide tests of unseen loading path simulations for both microstructures A and B. 
The training of the constitutive models used to make the forward predictions is described in Section~\ref{sec:constitutive_model_training} and the decoder used for each microstructure is described in Section~\ref{sec:autoencoder_training}.

\begin{figure}[h!]
\centering
\includegraphics[width=0.85\textwidth ,angle=0]{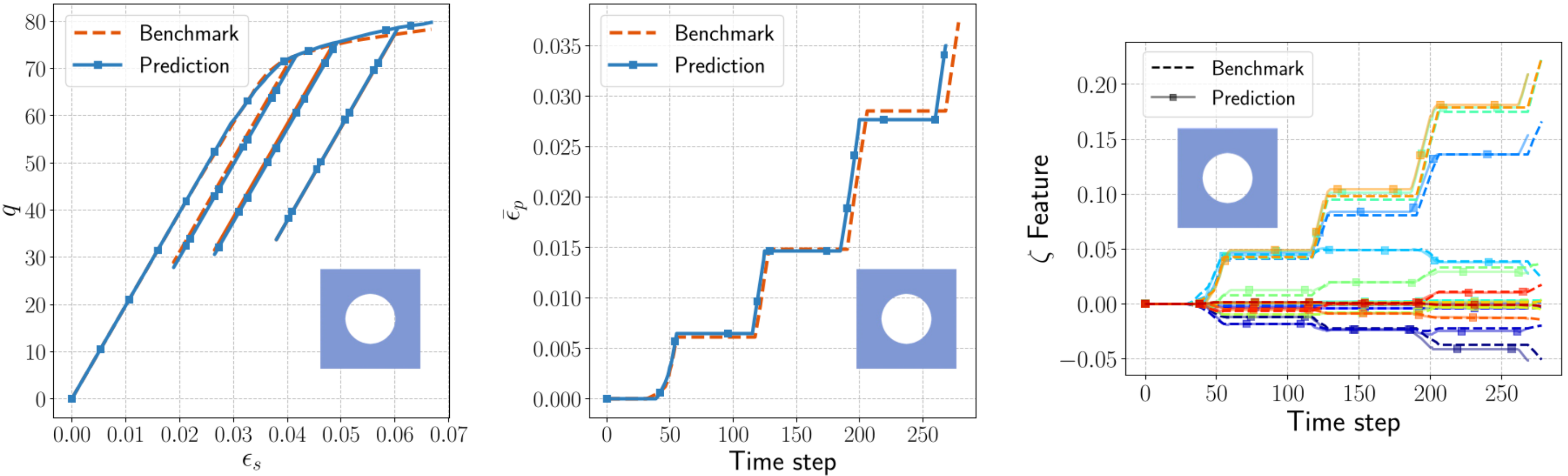} 
\caption{Prediction of deviatoric stress $q$, accumulated plastic strain $\overline{\epsilon}_p$, and the encoded feature vector $\tensor{\zeta}$ components using the return mapping algorithm for a cyclic loading of microstructure A.}
\label{fig:cyclic_loading_A}
\end{figure}

\begin{figure}[h!]
\centering
\includegraphics[width=0.85\textwidth ,angle=0]{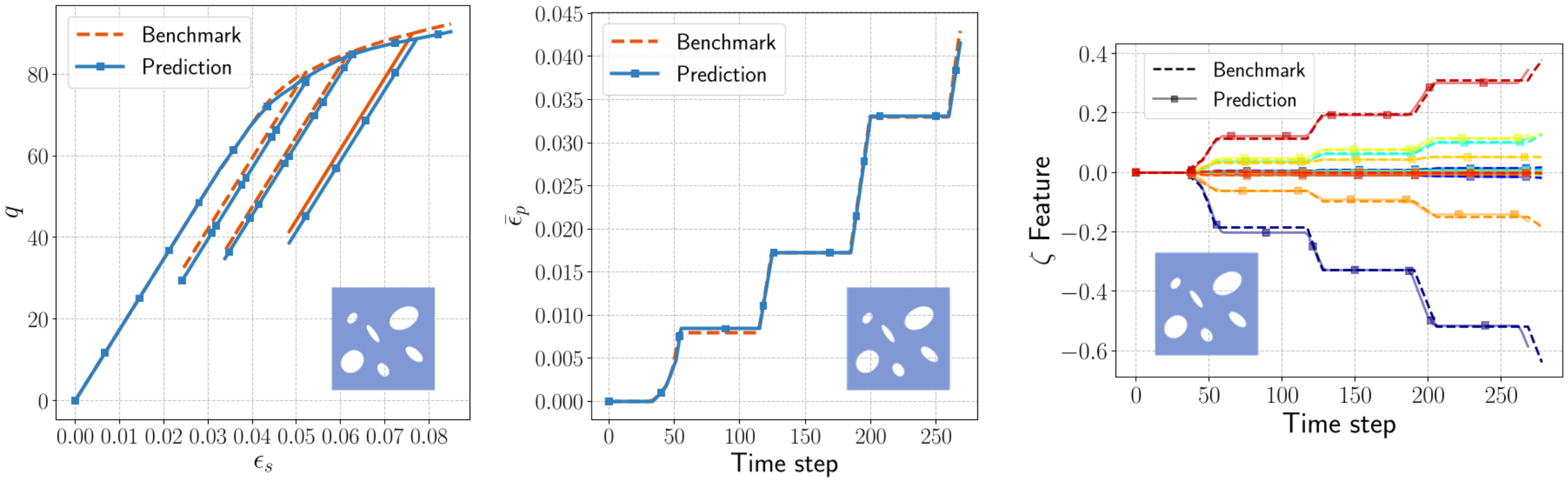} 

\caption{Prediction of deviatoric stress $q$, accumulated plastic strain $\overline{\epsilon}_p$, and the encoded feature vector $\tensor{\zeta}$ components using the return mapping algorithm for a cyclic loading of microstructure B.}
\label{fig:cyclic_loading_B}
\end{figure}

We first test the models' capacity to make predictions of the plastic state on monotonic data.
We demonstrate the result for the predicted stress state in Fig.~\ref{fig:monotonic_loading_A} (a \& b) and Fig.~\ref{fig:monotonic_loading_B} (a \& b) for microstructures A and B respectively.
We also record the homogenized plastic strain tensor of the microstructures. For simplicity, we are demonstrating the predicted accumulated plastic strain measure $\overline{\epsilon}_p$ in Fig.~\ref{fig:monotonic_loading_A} (c) and Fig.~\ref{fig:monotonic_loading_B} (c). Using the trained kinetic law neural network, we can make forward predictions of the encoded feature vectors $\widehat{\tensor{\zeta}}$ that are consistent with the current predicted homogenized plastic state. The results of these predictions are shown in \ref{fig:monotonic_loading_A} (d) and Fig.~\ref{fig:monotonic_loading_B} (d).
These predicted curves are a close match to the benchmark data and can closely capture the behavior in the macroscale. We can now interpret this homogenized behavior as the corresponding one in the microscale. Using the trained decoder for each microstructure, we can recover the plastic strain distributions as shown in Fig.\ref{fig:monotonic_loading_A} (e) and Fig.~\ref{fig:monotonic_loading_B} (e) for microstructures A and B respectively. It is noted that while only the node-wise prediction of the accumulated plastic strain $\overline{\epsilon}_p$ is shown, the decoded recovers the entire plastic strain tensor $\tensor{\epsilon}_p$. This is done for simplicity of presentation.
The node-wise predictions of the plastic strain are accurate and the decoder can qualitatively capture the general plastic distribution patterns and the plastic strain localization nodes in the microstructure.

We also demonstrate the capacity of the model to predict the plastic flow with the help of the encoded feature vector internal variables.
In Fig.~\ref{fig:flow_prediction}, we compare the computed plastic flow components using the neural network yield function stress gradient $\frac{\partial\widehat{f}}{\partial\sigma_A}$ prediction and the plastic potential stress gradient $\frac{\partial\widehat{\vartheta}}{\partial\sigma_A}$ as predicted by the plastic flow $\widehat{\tensor{g}}$ network with the benchmark simulations.
The results demonstrated are for two blind prediction curves for each microstructure -- Fig.~\ref{fig:flow_prediction}(a) corresponds to Fig.~\ref{fig:monotonic_loading_A} and Fig.~\ref{fig:flow_prediction}(c) corresponds to Fig.~\ref{fig:monotonic_loading_B}.
The accuracy of the $\widehat{\tensor{g}}$ network predictions on the flow is higher than that of the yield function stress gradient.
This is attributed to the decoupling of the yielding and hardening from the plastic flow directions allowing for more flexibility of the neural networks to fit these complex laws.
Network $\widehat{\tensor{g}}$ also utilizes the highly descriptive encoded feature vector $\tensor{\widehat{\zeta}}$ input that allows for more refined control of the plastic flow than the volume averaged accumulated plastic strain metric used in the yield function formulation. 

Finally, we conduct a similar blind test experiment but with added blind unloading and reloading elastic paths in the loading strains.
The results for microstructures A and B are demonstrated in Fig.~\ref{fig:cyclic_loading_A} and Fig.~\ref{fig:cyclic_loading_B} respectively.
As discussed in Section~\ref{sec:recurrent_architectures}, the model does not have any difficulty recognizing the elastic and plastic regions of the loading path since the behaviors are distinguished with the help of the neural network yield function.
This also constrains the evolution of the plastic strain and the encoded feature vector to happen only during the plastic loading.
Since the kinetic law neural network is a feed-forward architecture, there is no history dependence and no change in the plastic strain corresponds to no change in the encoded feature vector.
The decoder architecture is also path-independent so no change in the encoded feature vector corresponds to no change in the respective decoded plastic graph.
This is also achieved by the specific way the encoded feature vectors are constructed. The input node features in the autoencoder are the mesh node coordinates and the plastic strains as described in Section~\ref{sec:graph_representation}. This specific design ensures that the plastic graph does not evolve during elastic unloading/reloading and prevents any artificial memory effect in the elastic regime. 
The lack of memory effect in the elastic regime is necessary for the encoded feature vector to be internal variables for rate-independent plasticity models where 
the history-dependent effect is only triggered once the yield criterion is met. 
This would not be the case if other integration point data, such as the total strain or stress, are incorporated into the graph encoder. 

Note that this switch between path-independent and path-dependent behaviors may also have implications for other neural network constitutive laws. In particular,
if a black-box recurrent neural network is used to forecast history-dependent stress-strain responses, then one must ensure
that the history-dependent effect is not manifested in the elastic region. For instance, if the LSTM architecture is used, then one must ensure 
that the forget gate is trained to turn on to filter out any potential artificial influence of the strain history.

\section{Concluding Remarks}
\label{sec:conclucion}

The macroscopic inelastic behaviors of materials are manifested from the evolution of microstructures. As such, the major challenges 
of the macroscopic models used in engineering practice are just establishing sophisticated physics-informed predictions between input and 
output, but also exploring an efficient and economic way to represent the evolution of microstructures in a lower-dimensional space 
that is easier to establish models. In this work, we introduce a graph autoencoder framework that deduces internal variables 
that can effectively represent graph data obtained from representative elementary volumes. 
The encoder component of the autoencoder architecture maps data stored in a weighted graph onto a vector space spanned by low-dimensional descriptors that represents the key spatial features that govern the macroscopic responses, while 
the decoder component of the architecture projects these encoded feature vector-based descriptors for a robust interpretation of the material's behavior at the microscopic scale.
Through establishing this multiscale connection through the graph autoencoder, we introduce a new 
internal-variable driven plasticity framework where the long-lasting issues of the lack of interpretability of internal variables can be resolved.
A particularly important aspect of this framework is that allows us to bypass the process of hand-picking descriptors or geometric measures to establish plastic flow theory. This flexibility afforded by the autoencoder, therefore, creates accurate models, even though the optimal physical measures or descriptors of the mechanisms might not be known a priori. 
Note that, while the graphs may introduce a high-dimensional data structure, the macroscopic surrogate model is built in a low-dimensional $p-q$ space. Introducing a high-dimensional space for the yield function and the plastic flow (by, for instance, assuming no material symmetry or incorporating higher-order kinematics) may potentially lead to an even more accurate and precise model at the expense of increased difficulty in training and validation.
Another interesting aspect is to study the robustness of the proposed framework against different types of noise stored in different data structures, such as those stored in point sets (e.g. sensors distributed in three-dimensional objects), graphs (e.g. finite element solutions), and reconstructed manifolds.
Studies in these directions are currently in progress.

\section{Acknowledgments}
The authors are primarily supported by the National Science Foundation under grant contracts CMMI-1846875 and OAC-1940203, and
 the Dynamic Materials and Interactions Program from the Air Force Office of Scientific 
Research under grant contracts FA9550-19-1-0318,  FA9550-21-1-0391 and FA9550-21-1-0027, with additional 
support provided to WCS by the Department of Energy DE-NA0003962. 
These supports are gratefully acknowledged. 
The views and conclusions contained in this document are those of the authors, 
and should not be interpreted as representing the official policies, either expressed or implied, 
of the sponsors, including the Army Research Laboratory or the U.S. Government. 
The U.S. Government is authorized to reproduce and distribute reprints for 
Government purposes notwithstanding any copyright notation herein.

\bibliographystyle{plainnat}
\bibliography{main}

\end{document}